\documentclass[journal,12pt,onecolumn]{IEEEtran}

\usepackage{mathptmx}   
\usepackage{amsfonts, amssymb, amsmath}
\usepackage{epsfig}
\usepackage{theorem}
\usepackage{textcomp}
\usepackage{tipa}
\usepackage{multirow}
\usepackage{lipsum}

\makeatletter
\let\NAT@parse\undefined
\makeatletter
\usepackage{natbib}
\usepackage{hyperref}

\usepackage[linesnumbered,ruled]{algorithm2e}
\SetKwInput{KwIn}{Initial condition}
\SetKwInput{KwResult}{Outcome}

\newtheorem{theorem}{Theorem}

\newtheorem{corollary}[theorem]{Corollary}

\newtheorem{lemma}[theorem]{Lemma}
\newtheorem{problem}{Problem}

\newcommand{\qed}{\hfill $\Box$\\}

\def\argmin#1{\underset{#1}{\mathrm{argmin }}}
\def\argmax#1{\underset{#1}{\mathrm{argmax }}}

\newcommand{\EEE}{\mathbb{E}}
\newcommand{\ttr}{T_{\mathrm{tr}}}
\newcommand{\tobs}{T_{\mathrm{obs}}}
\newcommand{\lmax}{\lambda_{\mathrm{max}}}
\newcommand{\lmin}{\lambda_{\mathrm{min}}}

\begin{document}

\title{Persistent Monitoring of Events with Stochastic Arrivals at Multiple Stations}
\author{Jingjin Yu$^{1}$ \qquad\qquad Sertac Karaman$^{2}$ \qquad\qquad Daniela Rus$^{1}$
\thanks{$^{1}$Jingjin Yu and Deniela Rus are with the Computer Science and Artificial Intelligence Lab, Massachusetts Institute of Technology. E-mails: \{jingjin, rus\}@csail.mit.edu.}%
\thanks{$^{2}$Sertac Karaman is with the Department of Aeronautics and Astronautics, Massachusetts Institute of Technology. E-mail: sertac@mit.edu.}%
}
\date{ }
\maketitle

\begin{abstract}
This paper introduces a new mobile sensor scheduling problem, involving a single robot tasked with monitoring several events of interest that occur at different locations. Of particular interest  is the monitoring of transient events that can not be easily forecast. Application areas range from natural phenomena ({\em e.g.}, monitoring abnormal seismic activity around a volcano using a ground robot) to urban activities ({\em e.g.}, monitoring early formations of traffic congestion using an aerial robot). Motivated by those and many other examples, this paper focuses on problems in which the precise occurrence times of the events are unknown {\em a priori}, but statistics for their inter-arrival times are available. The robot's task is to monitor the events to optimize the following two objectives: {\em (i)} maximize the number of events observed and {\em (ii)} minimize the delay between two consecutive observations of events occurring at the same location. The paper considers the case when a robot is tasked with optimizing the event observations in a balanced manner, following a cyclic patrolling route. First, assuming the cyclic ordering of stations is known, we prove the existence and uniqueness of the optimal solution, and show that the optimal solution has desirable convergence and robustness properties. Our constructive proof also produces an efficient algorithm for computing the unique optimal solution with $O(n)$ time complexity, in which $n$ is the number of stations, with $O(\log n)$ time complexity for incrementally adding or removing stations. Except for the algorithm, most of the analysis remains valid when the cyclic order is unknown. We then provide a polynomial-time approximation scheme that gives a $(1+\epsilon)$-optimal solution for this more general, NP-hard problem. 
\end{abstract}

\section{Introduction}\label{section:introduction}
An avid documentary maker would like to observe several species of birds. Each species can be seen only at a particular location. Unfortunately, it is impossible to predict when exactly a bird will be seen at a sighting location. Hence, the documentary maker must wait in a hiding spot for the birds to appear. To the advantage of our documentary maker, past experience has furnished her with statistics of sighting times for each location. Given this information, the documentary maker would like to split her time between the locations, waiting to capture photos of bird sightings. While splitting her time, the documentary maker has two objectives. First, she would like to maximize the number of sightings. Second, she would like to minimize the delay between two consecutive sightings of the same species. Most importantly, our documentary maker is committed to striking a balance among the species. In other words, she would like to maximize the number of sightings and minimize the delay between two consecutive sightings, for all species all at the same time. 

\begin{figure}[t]
\begin{center}
\begin{tabular}{c}
    \includegraphics[width=3.4in]{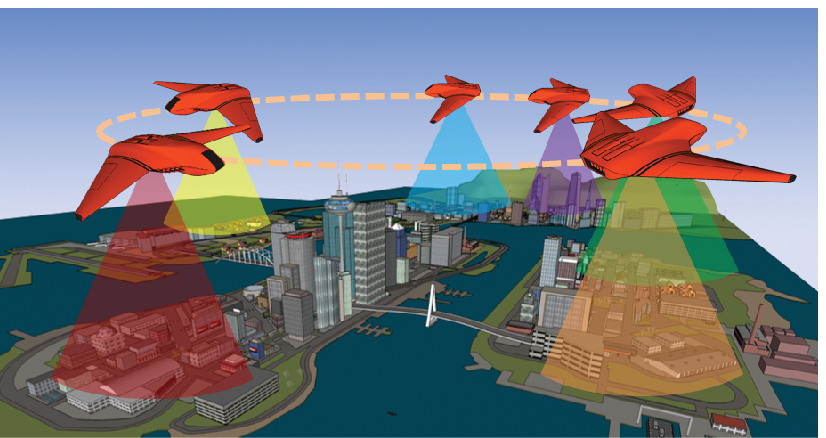} \\ (a) \\
    \includegraphics[width=3.4in]{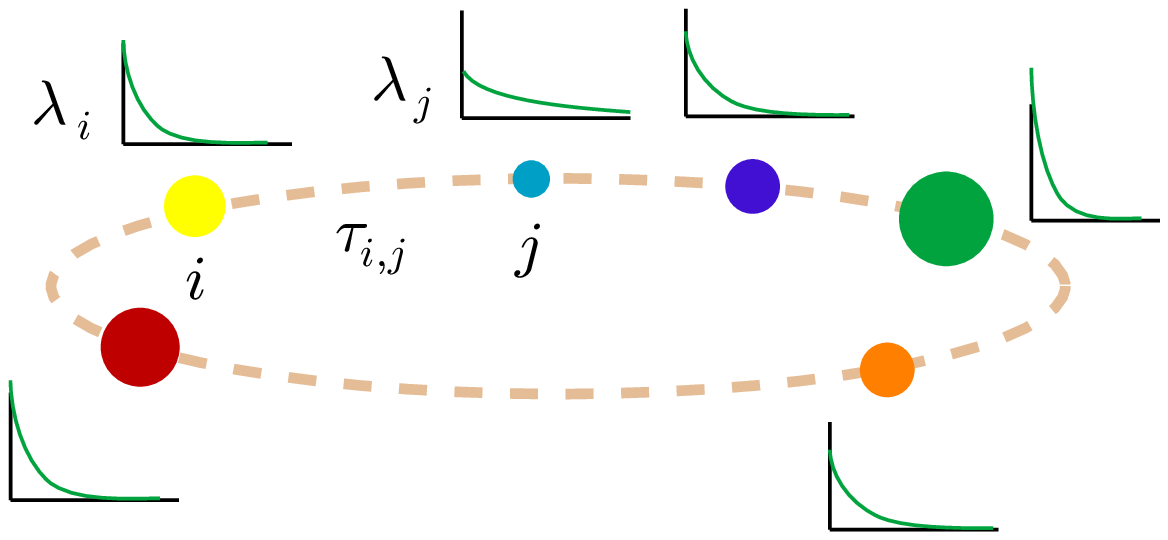} \\(b)
\end{tabular}
\end{center}
\caption{\label{fig:example} (a) One of many potential applications of our persistent monitoring formulation, in which an UAV (robot) is given the task of continuously observing randomly occurring (data) events at a set of fixed locations (the surface areas under the cones). The sizes of the discs represent the relative arrival rates of stochastic events at the locations. (b) Illustration of the underlying geometric problem setting. At each point of interest, say location (station) $i$, events arrive following a Poisson process with intensity $\lambda_i$. It takes a robot $\tau_{i,j}$ time to move from station $i$ to station $j$, during which no observation can be made. The associated plots roughly capture the (exponential) distributions of event arrivals associated with the stations.}
\end{figure}

The bird documentary maker example captures the essential elements of the problem studied in this paper. More formally, consider a single robotic vehicle tasked with monitoring stochastic and transient events that occur at multiple locations (see, {\em e.g.}, Figure~\ref{fig:example}). Unable to predict exactly when an event happens, the robot must travel to a particular location and wait for the event to occur. Limited by a single robot, its schedule\footnote{In this paper, {\em schedule} and {\em policy} are used interchangeably.} must be optimized to ensure that all locations are observed equally well as best as possible, {\em i.e.}, in a balanced manner, according to the following objectives: {\em (i)} ensure that a large number of events are observed at each location and {\em (ii)} ensure that the delay between two observations of events at any given location is minimized. Optimizing these objectives in a balanced manner gives rise to a multi-objective mobile sensor scheduling problem. This paper is concerned with the mathematical analyses and algorithmic approaches for this complex multi-objective optimization problem.


The problem we study is applicable to a broad set of practical scenarios, including surveillance and reconnaissance, and scientific monitoring. The events of interest include natural phenomena ({\em e.g.}, volcanic eruptions and early formations of blizzards, hailstorms, and tsunamis), biological disasters ({\em e.g.}, early formations of epidemic diseases on animal or plant populations), military operations ({\em e.g.}, terrorist attacks), among others. The key common characteristic of these events is that their precise time of occurrence can not be easily forecast, although the statistics regarding how often they occur may be available, for example, from past experience. Hence, the data-collecting robot must wait at the location of interest to capture the event once it occurs. Then, the fundamental scheduling problem is to decide how much time the robot should spend at each location to achieve various objectives, such as those described above. Our main theoretical result is that this complex multi-objective mobile sensor scheduling problem can be reduced to a quasi-convex optimization problem, which implies efficient algorithms for computing optimal solutions. In particular, globally optimal or near-optimal solutions can be computed in time polynomial in the number of locations.

{\bf Related Work:}
Broadly speaking, persistent monitoring problems appear naturally whenever only limited resources are available for serving a set of spatially-dispersed tasks. Motivated by a variety of potential applications, such as aerial \cite{MicStuMoh11} and underwater \cite{SmiSchSmiJonRusSuk11} data collection, several authors have studied persistent monitoring problems~\cite{AlaFatSmi12,ArvKimMar12,CasLinDing13,GirHowHed05,GroKelKumPap06,LanSchwagerICRA13GRFPersistentMonitoring,NigKro05,SmithSchwagerRusTRO12Persistence,SolteroIROS12PathMorphing}. In \cite{AlaFatSmi12}, the authors consider a weighted latency measure as a robot continuously traverse a graph, in which the vertices represent the regions of interest and the edges between the vertices are labeled with the travel time. They present a $O(\log n)$-approximation algorithm for the proposed problem. In \cite{ArvKimMar12}, a memoryless control policy is designed to guide robots modeled as controllable Markov chains to maximize their monitoring area while avoiding hazardous areas. In \cite{CasLinDing13}, the authors consider a persistent monitoring problem for a group of agents in a 1D mission space. They show that this problem can be solved by parametrically optimizing a sequence of switching locations for the agents. 

The coordination and surveillance problem for multiple unmanned aerial vehicles is addressed in \cite{GirHowHed05,NigKro05}. Coordination among aerial and ground vehicles are further explored in \cite{GroKelKumPap06}. A random sampling method that generates optimal cyclic trajectories for monitoring Gaussian random field is presented in \cite{LanSchwagerICRA13GRFPersistentMonitoring}. The problem of generating speed profiles for robots along predetermined closed paths for keeping bounded a varying field is addressed in \cite{SmithSchwagerRusTRO12Persistence}. The authors characterized policies for both single and multiple robots. In \cite{SolteroIROS12PathMorphing}, decentralized adaptive controllers were designed to morph the initial closed paths of robots to focus on regions of high importance. 

In contrast to the references cited above, the problem studied in this paper focuses on transient events at discrete locations, emphasizing unknown arrival times (but known statistics). Since an event is only observable at discrete locations, the event arrival times being unknown forces the robot to wait at each station in order to observe the events of interest. Waiting at a station then introduces delay at other stations. This stochastic event model links our work to stochastic vehicle routing problems such as the dynamic traveling repairman problem (DTRP) that is studied in \cite{BerRyz93,PavFraBul11}, among others. Our problem differs from problems like DTRP in that we are not concerned with capturing all events, but rather collecting a reasonable amount of events across the locations in a balanced manner without large gaps between observations of events from the same location. 

Persistent surveillance problems are intimately connected to coverage problems. Coverage of a two-dimensional region has been extensively studied in robotics \cite{Cho00,Cho01,GabRim03}, as well as in purely geometric settings. For example, in~ \cite{ChiNta88}, the proposed algorithms compute the shortest closed routes for continuous coverage of polygonal interiors under an infinite visibility sensing model. Coverage with limited sensing range was also addressed \cite{HokStiSpo08,Nta91}. If the environment to be monitored has a 1-dimensional structure, discrete optimization problems, such as the traveling salesmen problem (TSP), often arise \cite{AlaFatSmi12}. 
In most coverage problems, including those cited above, the objective is to place sensors in order to maximize, for example, the area that is within their sensing region. The persistent surveillance problem we study in this paper is a special case, in which limited number of sensors do not allow extensive coverage; hence, we resort to mobility in order to optimize the aforementioned performance metrics. 

Persistent monitoring problems are also related to (static) {\em sensor scheduling} problems (see, {\em e.g.},~\cite{FueVee08,HeChong04,HerKreBla08}), which are usually concerned with scheduling the activation times of sensors in order to maximize the information collected about a time-varying process. 
The problem considered in this paper involves a {\em mobile sensor} that can travel to each of the locations, in which the additional time required to travel between stations is non-zero.
The mobile sensor scheduling literature is also rich. 
For instance, in~\cite{LeDahFerFra08}, the authors study the control of a robotic vehicle in order to maximize data rate while collecting data stochastically arriving at two locations. 
The problem studied in this paper is a novel mobile sensor scheduling problem involving several locations and a multi-objective performance metric that includes both the data rate and the delay between consecutive observations.

{\bf Contributions:}
The contributions of this paper can be summarized as follows. First, we propose a novel persistent monitoring and data collection problem, with the unique feature that the precise arrival times of events are unknown {\em a priori}, but their statistics are available. Combined with the assumption that the events are generated at distributed, discrete locations, the stochastic event model allows our formulation to encompass many practical applications in which the precise occurrence times of the events of interest can not be forecast easily. Second, we prove that this fairly complex multi-objective mobile sensor scheduling problem admits a unique, globally optimal solution in all but rare degenerate cases. The optimal solution is also shown to have desirable convergence property and robustness. Moreover, the unique policy can be computed extremely efficiently when the station visiting order is predetermined and efficiently to $(1+\epsilon)$-optimal when the visiting order is not given {\em a priori}. At the core of our analysis is a key intermediate result that reduces the mobile sensor scheduling problem to a quasi-convex optimization problem in one variable, which may be of independent interest.

This paper builds on \cite{YuKarRus14ICRA-A} and significantly extends the conference publication in the following aspects: {\em (i)} in addition to existence, solution uniqueness is now established, {\em (ii)} convergence and robustness results are introduced and thoroughly discussed to render the study more complete,  {\em (iii)} a polynomial-time approximation algorithm is provided that solves the more general problem when the cyclic ordering of stations is unknown {\em a priori}, and {\em (iv)} extensive computational experiments are added to confirm our theoretical development as well as to provide insights into the structure of our proposed optimization problem. 

The rest of the paper is organized as follows. In Section~\ref{section:problem}, we provide a precise definition of the multi-objective persistent monitoring problem that we study. Starting with the assumption that the stations' cyclic order is known, we prove existence and uniqueness of optimal solutions to this slightly restricted problem in Section~\ref{section:theory}. We further explore the convergence and robustness properties of the optimal solution in Section~\ref{section:property}. In Section~\ref{section:algorithm}, we deliver algorithmic solutions for the multi-objective optimization problem with and without a predetermined station visiting order, and characterize their computational complexity. We present and discuss computational experiments in Section~\ref{section:experiment}, and conclude the paper in Section~\ref{section:conclusion}. Frequently used symbols are listed in Table~\ref{table_symbols}.

\begin{table}[t]
\begin{center}
	\caption{\label{table_symbols} List of frequently used symbols and their interpretations. }
	\begin{tabular}{cl}
	 \hline\hline \\
	 $\lambda_i$ &  Arrival rate of the Poisson process at station $i$\\
	 $\tau_{i,j}$ & Travel time from station $i$ to station $j$\\	 
	 $\pi$ & \parbox[t]{6.5cm}{Cyclic policy of the form $((k_1, t_1), \dots, $ $(k_n,t_n))$, in which $t_i$ is the time spent by the robot at station $k_i$, $1 \le k_i \le n$, in one policy cycle, or of the form $(t_1, \ldots, t_n)$ when $k_i = i$} \\
	 $J_i(\pi)$ & An objective function to be optimized\\
	 $T$ & Total time incurred by a policy cycle\\
	 $\ttr$ & Total travel time per policy cycle\\
	 $\tobs$ & $T - \ttr$, total observation time per policy cycle \\
	 $\sigma$ & $1/(\sum_{i = 1}^n(1/\lambda_i))$, the harmonic sum of $\lambda_i$'s\\
	 $\gamma_i$ & $\sigma/\lambda_i = 1/(\lambda_i\sum_{j = 1}^n(1/\lambda_j))$\\
	 $N_i(\pi)$ & \parbox[t]{6.5cm}{The number of events collected at station $i$ in one period of the policy $\pi$}\\
	 $T_i(\pi)$ & \parbox[t]{6.5cm}{The time between two consecutive event observations at station $i$ containing travel to other stations, for the policy $\pi$} \\
	 $Pr(e)$ & \parbox[t]{6.5cm}{Probability of an event $e$}\\
	 $\EEE[X]$ & Expected value of a random variable $X$\vspace{2mm}\\
	 $\alpha_i(\pi)$ & $\EEE[N_i (\pi)]/ \sum_{j = 1}^n \EEE[N_j (\pi)]$\vspace{2mm}\\
	 $\Delta_{ij}(\pi)$ & $|\EEE[N_i(\pi)] - \EEE[N_j(\pi)]|$\\
	 $\Pi$ & $\arg\max_{\pi} \min_{i}\alpha_i (\pi)$\\
	 \hline\hline
	 \end{tabular}
\end{center}
\end{table}
\section{Problem Statement} \label{section:problem}

Consider a network of $n$ {\em stations} or {\em sites} that are spatially distributed in $\mathbb R^2$. At each station, interesting but {\em transient} events may occur at unpredictable time instances. The arrival times of events at a station $i$, $1 \le i \le n$, are assumed to follow a Poisson process with a known (mean) {\em arrival rate} or {\em intensity} $\lambda_i$, with a unit of number of events per hour. The event arrival processes are assumed to be independent between two different stations. Let there be a mobile robot that travels from station to station. The robot is equipped with on-board sensors, such as cameras, that allow the robot to record data containing the stochastic events occurring at the stations. Let $\tau_{i,j}$ denote the time it takes the robot to travel from station $i$ to station $j$. We assume that $\tau_{i,j}$ is proportional to the Euclidean distances between stations $i$ and $j$. 

We want to design {\em cyclic policies} to enable optimal data collection, according to the objectives described in the introduction. A precise definition of these objectives will follow shortly. In a cyclic policy, the robot visits the stations in a fixed (but unknown {\em a priori}) cyclic order and wait at each station for a fixed amount of time to collect data. The solution scheduling policy then takes the form $\pi = ((k_1, t_1), \ldots, (k_n, t_n))$ in which $k_i$'s describe the visiting order and $t_i$'s describe the waiting time of the robot at station $k_i$. 

\textbf{Remark.} Having fixed waiting time suggests that the policy is an open-loop ({\em i.e.}, no feedback) policy. We note that such policies are of practical importance. For example, it may be the case that an aerial mobile robot only has limited energy or computing power to process the data ({\em e.g.}, a large number of video streams) it collects. Similarly, an underwater robot gathering plankton samples may not have on-board equipment to analyze the collected samples. As another example, the transportation of the data collection equipments can be a non-trivial task which requires that the travel schedule to be prearranged. In yet another example, in certain scenarios, it may even be desirable not to allow the robot to have immediate semantic understanding of the collected data due to security reasons such as hacking prevention. 

Given a policy $\pi$, we define its {\em period} as
\[
T := \sum_{i = 1}^{n -1}\tau_{k_i,k_{i+1}} + \tau_{k_n,k_1} + \sum_{i = 1}^n t_i.
\]
For convenience, let $\ttr := \sum_{i = 1}^{n -1}\tau_{k_i,k_{i+1}} + \tau_{k_n,k_1}$ be the total travel time per policy cycle and $\tobs := \sum_{i = 1}^n t_i $ $= T - \ttr$ be the total observation time per policy cycle. Let $N_i (\pi)$ denote the number of events observed at station $k_i$ in one cycle. For the first objective, seeking to ensure maximal and equal priorities are allocated to all stations, we maximize the {\em fraction of events observed at each station in a balanced manner}, {\em i.e.}, 

\begin{align} \label{equation:objective_one}
J_1(\pi) = \min_{i}\,\, \alpha_i(\pi) = \min_{i}\,\, \frac{\EEE[N_i(\pi)]}{\sum_{j = 1}^n \EEE[N_j(\pi)]},
\end{align}
subject to the additional constraint 
\begin{align} \label{equation:objective_one_constraint}
\pi \in \argmin{\pi} \max_{i,j}\Delta_{ij}(\pi) = \argmin{\pi}\max_{i,j}|\EEE[N_i(\pi)] - \EEE[N_j(\pi)]|,
\end{align}
which further balances event observation efforts by penalizing large observation discrepancies between different stations. Alternatively, one may view~\eqref{equation:objective_one_constraint} as a higher order balancing effort than maximizing $J_1$. 

\begin{figure}[htp]
\begin{center}
\includegraphics[width=0.48\textwidth]{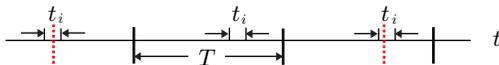} 
\end{center}
\caption{\label{fig:delay} Illustration of an {\em instance} or a {\em sample} of a {\em delay}. The policy has a period of $T$ and the robot is waiting at station $k_i$ during the intervals with length $t_i$. The two dotted lines correspond to times when events occur at station $k_i$. If no other events happen at station $k_i$ in between these two times when the robot is present at station $k_i$, then the time between these two event occurrences is an instance of the delay at station $k_i$.}
\end{figure}

The second objective seeks to minimize large delays between event observations at the same station. We formalize the notion of {\em delay}, a random variable, as follows. As the robot executes a policy, it arrives at station $k_i$ periodically and waits for $t_i$ time at station $k_i$ each time it gets there. Suppose that during one such waiting time $t_i$, one or more events occur at station $k_i$. Let the last event within this particular $t_i$ be $t_{start}$. An instance of a delay is a period of time that begins at $t_{start}$ and ends when another event occurs at station $k_i$ while the robot is waiting at station $k_i$ (see, {\em e.g.}, Figure~\ref{fig:delay}). More precisely, we define the delay at station $k_i$ for a given policy $\pi$, denoted $T_i(\pi)$, as a random variable that maps these instances of delays to probability densities. Our second objective aims at minimizing the {\em maximum delay across all stations}, {\em i.e.}, 
\begin{align} \label{equation:objective_two}
J_2(\pi) = \max_{i} \,\,\EEE[T_i (\pi)].
\end{align}

Often, both objectives are equally important. One would like to spend as much time as possible at all stations for maximizing the data collection effort and at the same time minimize delays between observations at any given station, which is penalized if the robot lingers at any station for too long. Interestingly, the set of policies that optimizes the first objective function is not unique; in fact, there are infinitely many such policies. We compute the optimal policy for the second objective function among those policies that optimize the first objective function. That is, we compute the policy
$
\pi^* = \arg\min_{\pi \in \Pi} J_2(\pi),
$
with 
$
\Pi := \arg\max_{\pi'} J_1(\pi'),
$
subject to~\eqref{equation:objective_one_constraint}. We then further show that $\pi^*$ is the {\em unique} Pareto optimal solution for optimizing both $J_1$ and $J_2$. 

\vspace*{1mm}
With the setup so far, we now formally state the persistent monitoring problem studied in this paper.
\vspace*{-2mm}
\begin{problem}\label{problem:general}
Given $n$, $\{\lambda_i\}$, and $\{\tau_{i,j}\}$, find an optimal solution $\pi^* = ((k_1^*, t_1^*), \ldots, (k_n^*, t_n^*))$ that optimizes $J_1$ and $J_2$ subject to the constraint equation~\eqref{equation:objective_one_constraint}.
\end{problem}

To facilitate our analysis, we begin with a special case in which the visiting order of stations are fixed {\em a priori}. That is, we assume $k_i = i$. 
\begin{problem}\label{problem:cycle}Given $n$, $\{\lambda_i\}$, and $\{\tau_{i,j}\}$, and assume that the robot visits the $n$ stations in the cyclic order of $1, 2, \ldots, n$, find an optimal solution $\pi^* = (t_1^*, \ldots, t_n^*)$ that optimizes $J_1$ and $J_2$ subject to the constraint equation~\eqref{equation:objective_one_constraint}.
\end{problem}

It is straightforward to observe that the travel times only matter as a whole, {\em i.e.}, the policy's dependency on the robot's path in a cyclic policy only hinges on $\ttr$. The immediate gain from proposing Problem~\ref{problem:cycle} is that it removes the need to compute a TSP tour, allowing us to focus our study on the optimality structure induced by $J_1$ and $J_2$, which is fairly rich. We dedicate Sections~\ref{section:theory} and~\ref{section:property} to Problem~\ref{problem:cycle} and revisit Problem~\ref{problem:general} when we discuss algorithmic solutions in Section~\ref{section:algorithm}.



\section{Existence and Uniqueness of Optimal Policy}\label{section:theory}
In this section, we establish the existence and uniqueness of solutions for Problem~\ref{problem:cycle}. We also simply refer to a policy $\pi$ as $\pi = (t_1, \ldots, t_n)$ here and in Section~\ref{section:property}.  We note that the results from this section go beyond simply showing the existence and uniqueness of an optimal cyclic policy for the robot; an effective means for computing such a policy is also implied. The discussion of the implied algorithmic solution is deferred to Section~\ref{section:algorithm}.

\subsection{Existence of Optimal Solution}
We now establish the existence of optimal solutions to Problem~\ref{problem:cycle}. Showing the existence of solution to a multi-objective optimization problem requires showing that the Pareto front is not empty. We achieve this goal through Theorem~\ref{theorem:optimality}, which works with the two objectives sequentially. Theorem~\ref{theorem:optimality} does more than simply showing the Pareto front is non-empty; it actually describes an optimization program that finds a point on the Pareto front. 

\begin{theorem}[Existence of Optimal Solution]\label{theorem:optimality}
There exists a continuum of policies that maximize $J_1$ under the constraint~\eqref{equation:objective_one_constraint}, given by
$$
\Pi := \argmax{\pi} \min_{i}\,\, \frac{\EEE[N_i(\pi)]}{\sum_{j = 1}^n \EEE[N_j(\pi)]}.
$$
Among all policies in $\Pi$, there is a unique policy $\pi^*$ that minimizes $J_2$. Moreover, this unique policy $\pi^* = (t_1^*, t_2^*, \dots, t_n^*)$ is determined by
$$
t_i^* = \frac{\sigma}{\lambda_i} \tobs^*,
$$
in which
$$
\tobs^* := \argmin{\tobs} \left(\frac{2}{\lmax} + \frac{(\tobs + \ttr) \lmax - \sigma \tobs (1+ e^{- \sigma \tobs})}{(1 - e^{- \sigma \tobs})\lmax} \right),
$$
with $\lmax = \max_{i} \lambda_i$ being the maximum arrival rate and $\sigma = \left( \sum_{i = 1}^n \lambda_i^{-1} \right)^{-1}$ the harmonic sum of $\lambda_i$'s.\footnote{Harmonic mean is usually defined as $\lambda_\mathrm{hm}= \left((1/n) \sum_{i=1}^n \lambda_i^{-1} \right)^{-1}$. Accordingly, we define the harmonic sum as $\sigma = n \, \lambda_{hm}$.} The optimization problem that gives $\tobs^*$ is a quasi-convex program in one variable, the unique optimal solution for which can be computed efficiently, for example, by using the Newton-Raphson method to compute the root of the derivative of its objective function.
\end{theorem}

To prove Theorem~\ref{theorem:optimality}, we need several intermediate results, which are stated and proved through Lemmas~\ref{lemma_continuum}-\ref{lemma:monotonic_lambda}. Our constructive proof of Theorem~\ref{theorem:optimality} begins by characterizing policies that maximize the first objective.
\begin{lemma}\label{lemma_continuum}
%
Among all cyclic policies, a cyclic policy $\pi$ maximizes $J_1(\pi)$ under the constraint~\eqref{equation:objective_one_constraint}, for any $\tobs > 0$, if and only if
\begin{equation}\label{eq:sol1}
t_i = \displaystyle \frac{\sigma \tobs}{\lambda_i} = \frac{\tobs}{\lambda_i\sum_{j=1}^n\frac{1}{\lambda_j}}.
\end{equation} 
Moreover, such a cyclic policy $\pi$ satisfies:
\begin{equation}\label{eq:eq}
\EEE[N_1 (\pi)] \,\,=\,\, \EEE[N_2(\pi)] \,\,=\,\, \cdots \,\,=\,\, \EEE[N_n(\pi)].
\end{equation}
\end{lemma}

\noindent{\sc Proof.} By linearity of expectation, the value of $J_1$, as defined in (\ref{equation:objective_one}), remains the same if we only look at a single policy cycle. We show that for arbitrary $\tobs > 0$, choosing $t_i$'s according to (\ref{eq:sol1}) yields the same optimal value for $J_1$. Now fixing a policy $\pi$, after spending $t_i$ time at station $i$, the robot collects $\EEE[N_i(\pi)] = \lambda_i t_i$ data points in expectation. This yields 
$$
\alpha_i(\pi) = \frac{\EEE[N_i(\pi)]}{\sum_{j=1}^n\EEE[N_j(\pi)]} = \frac{\lambda_i t_i}{\sum_{j=1}^n \lambda_j t_j}.
$$

By the pigeonhole principle, $\min_{i}\alpha_i(\pi)$ is maximized if and only if~\eqref{eq:eq} is satisfied, yielding $J_1 = 1/n$. When~\eqref{eq:eq} holds, the constraint~\eqref{equation:objective_one_constraint} is satisfied since it achieves a value of zero. Solving the set of equations  
$$
\left\{\begin{array}{l}\lambda_1t_1 = \ldots = \lambda_nt_n\\ 
\displaystyle\sum_{i = 1}^n t_i = \tobs \end{array}\right.
$$
then yields~\eqref{eq:sol1}. ~\qed

\textbf{Remark.} Lemma~\ref{lemma_continuum} implies that any cyclic policy that equalizes $\EEE[N_i(\pi)]$ across the stations optimizes the first objective $J_1$. This provides us with an infinite set of policies that are optimal for the first objective function. Any policy satisfying~\eqref{eq:sol1} is optimal, independent of the value of the policy period $T$. 

Next, we show that, among the set of policies $\Pi$ provided by~Lemma~\ref{lemma_continuum}, there exists a unique $\pi^*$ that optimizes the second objective $J_2$. To achieve this, a method for evaluating $\EEE[T_i(\pi)]$ is required. It turns out that an analytical formula can be derived for computing $\EEE[T_i(\pi)]$. 

\begin{figure}[htp]
\begin{center}
\includegraphics[width=0.48\textwidth]{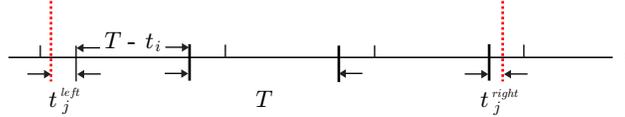} 
\end{center}
\caption{\label{figure_observation_gap} Illustration of the components of an observation gap.}
\end{figure}

\begin{lemma}\label{l:exp_break_cyclic}
Let $\pi = (t_1, \ldots, t_n)$ be a cyclic policy and let $T = \ttr  + \sum_{i = 1}^nt_i$ be the period of the cyclic policy. Then 
\begin{equation}\label{eq:exp_break_cyclic}
\EEE[T_i(\pi)] = \displaystyle\displaystyle\frac{2}{\lambda_i} +  \frac{T - t_i - t_ie^{-\lambda_it_i}}{1 - e^{-\lambda_it_i}}.
\end{equation}
\end{lemma}
\noindent{\sc Proof.} To compute $\EEE[T_i(\pi)]$, without loss of generality, fix an observation window at station $i$ and call it observation window $0$, or $o_0$. We may further assume without loss of generality that $o_0$ contains the arrival of at least one event at station $i$. We look at all observation gaps on the right of $o_0$. The left side of $o_0$ may be safely ignored due to the memoryless property of Poisson processes. Any observation gap $g_j$ contains the following parts, from left to right: 1. $t_j^{left}$, the overlap of $g_j$ with the observation window on $g_j$'s left end, 2. $T - t_i$, the first observation break (an observation break for station $i$ is the time window between two consecutive visits to station $i$), 3. $0 \le m < \infty$ additional policy cycles (of length $T$ each), and 4. $t_j^{right}$, the overlap of $g_j$ with the observation window on $g_j$'s right end. As an example, in Figure~\ref{figure_observation_gap}, the start and end of the observation gap $g_j$ are marked with the two  dotted lines. The parts $t_j^{left}$, the first observation break $T-t_i$, and $t_j^{right}$ are also marked. The gap $g_j$ further contains two additional policy cycles, {\em i.e.}, $m = 2$. 

The computation of $\EEE[T_i(\pi)]$ is a two-step process: 1. compute the probability $p_m$ of a gap $g_j$ spanning up to $m+2$ policy cycles for any $m \ge 0$, and 2. compute $\EEE[T_i(\pi)]$ as
\begin{equation}\label{equation_eeti_expression}
\EEE[T_i(\pi)] = \sum_{m = 0}^{\infty} p_m \EEE_m,
\end{equation}
in which $\EEE_m$ is the expected length of the gap $g_j$. Note that~\eqref{equation_eeti_expression} holds as long as the expectations $\EEE[T_i(\pi)]$ and $\EEE_m$ are over the same underlying distribution. We compute $\EEE_m$ with 
\begin{equation}\label{equation:tipi}
\begin{array}{l}
\displaystyle \EEE_m = \EEE[t_j^{left}] + \EEE[t_j^{right}] + T - t_i + mT\\
\displaystyle \quad = \displaystyle 2\EEE[t_j^{left}] + T - t_i + mT.
\end{array}
\end{equation}

A time reversed Poisson process is again a Poisson process with the same arrival rate. Due to this symmetry along the time time, the second equality in~\eqref{equation:tipi} holds because $\EEE[t_j^{left}] = \EEE[t_j^{right}]$. To compute $p_m$, note that we never need to consider the left side of a gap $g_j$. This is true because as we look at an infinite sequence of consecutive gaps $g_1, \ldots, g_j, \ldots$. The left most observation window (which is $o_0$) overlapping with $g_1$ is fixed by assumption. Once the right most observation window overlapping with $g_1$ is set (with certain probability), this explicitly fixes the left most observation window overlapping with $g_2$ and recursively, the left most observation window overlapping $g_j$. Therefore, the probability of $g_j$ spanning $m + 1$ policy cycles is 
\begin{align}\label{equation:pm}
p_m = e^{-m\lambda_it_i}(1-e^{-\lambda_it_i}). 
\end{align}

The first term of~\eqref{equation:pm}, $e^{-m\lambda_it_i}$, is the probability that $g_j$ contains $0, 1, \ldots, m-1$ full policy cycles. The probability of no event happening in each additional cycle in the sequence is $e^{-\lambda_it_i}$. They can be combined because the exponential distribution is memoryless. The term $(1-e^{-\lambda_it_i})$ is the probability that at least one event happens in the right most observation window overlapping $g_j$. Noting that the terms $2\EEE[t_j^{left}] + T - t_i$ appear in all $\EEE_m$'s, we can rewrite $\EEE[T_i(\pi)]$ as 
\begin{equation}\label{equation_eeti_expression_2}
\begin{array}{l}
\EEE[T_i(\pi)] = \displaystyle 2\EEE[t_j^{left}] + T - t_i + \sum_{m = 1}^{\infty} mTe^{-m\lambda_it_i}(1-e^{-\lambda_it_i})\\
\end{array}
\end{equation}
in which 
\begin{equation}\label{equation_additional_cycles}
\begin{array}{l}
\displaystyle\sum_{m = 0}^{\infty} mTe^{-m\lambda_it_i}(1-e^{-\lambda_it_i})  = \displaystyle T(1-e^{-\lambda_it_i})\sum_{m = 1}^{\infty} \sum_{k = m}^{\infty} e^{-k\lambda_it_i} \\
\quad = \displaystyle T(1-e^{-\lambda_it_i})\sum_{m = 1}^{\infty} \frac{e^{-m\lambda_it_i}}{1-e^{-\lambda_it_i}}  = \displaystyle \frac{Te^{-\lambda_it_i}}{1-e^{-\lambda_it_i}}.
\end{array}
\end{equation}

The computation of $\EEE[t_j^{left}]$ is carried out as follows. By assumption, at least one event happens during the given observation window of length $t_i$. Let the number of events within this $t_i$ time be $N_e$. The probability of having $k$ events is $Pr(N_e = k) = (\lambda_it_i)^ke^{-\lambda_it_i}/k!$. Let $\tau_{1, k}$ be the arrival time of the first event among $k$ events. For each $k \ge 1$, the distribution of the $k$ events is uniform over $[0, t_i]$. We have for $0 \le t \le t_i$, 
$$
Pr(\tau_{1, k} > t) = \big(\frac{t_i - t}{t_i}\big)^k, 
$$
from which we can obtain the probability density function for $\tau_{1, k}$ and then $\EEE[\tau_{1, k}] = t_i/(k+1)$. Then
\begin{equation}\label{equation_e_left}
\begin{array}{l}
\displaystyle\EEE[t_j^{left}] = \frac{\displaystyle\sum_{k = 1}^{\infty}\EEE[\tau_{1,k}]Pr(N_e = k) }{1 - Pr(N_e = 0)} = \displaystyle\frac{\displaystyle\sum_{k = 1}^{\infty}\frac{t_i}{k + 1}(\lambda_it_i)^ke^{-\lambda_it_i}/k! }{1 - e^{-\lambda_it_i}}\\
\displaystyle \quad = \frac{1}{1 - e^{-\lambda_it_i}} \sum_{k = 1}^{\infty}\frac{t_i(\lambda_it_i)^ke^{-\lambda_it_i}}{(k + 1)!} \\
\displaystyle \quad = \displaystyle\frac{1}{\lambda_i(1 - e^{-\lambda_it_i})}(1 - e^{-\lambda_it_i} - \lambda_it_ie^{-\lambda_it_i}) = \frac{1}{\lambda_i} - \frac{t_i e^{-\lambda_it_i}}{1 - e^{-\lambda_it_i}}.
\end{array}
\end{equation}
Finally, plugging~\eqref{equation_additional_cycles} and~\eqref{equation_e_left} into~\eqref{equation_eeti_expression_2} yields~\eqref{eq:exp_break_cyclic}. ~\qed

\textbf{Remark.} The technique from Lemma~\ref{l:exp_break_cyclic} is generic and can be used to compute expectations of other types of delays. For example, the current $\EEE[T_i(\pi)]$ treats delays with different values of $m$ with equal importance. It may be the case that we want to further penalize for not observing any events over longer periods of time. One simple way to enable this is to give weights to delays with larger $m$ values. This can be incorporated easily by updating $\EEE_m$ to 
$$
\displaystyle \EEE_m = \EEE[t_j^{left}] + \EEE[t_j^{right}] + T - t_i + m^2T.
$$

The remaining steps for computing this alternative expected delay stay unchanged. 

With $\EEE[T_i(\pi)]$ for each of the $1 \le i \le n$ stations, finding the optimal policy among $\Pi$ that minimizes $J_2$ remains a nontrivial task. To obtain $\min_{\pi} \max_{i}\EEE[T_i(\pi)]$, we have to build the upper envelope over $n$ such expected delays and then locate the minimum on that envelope.  Fortunately, $\EEE[T_i(\pi)]$ has some additional properties that make this task more manageable. One such property is that $\EEE[T_i(\pi)]$ is quasi-convex in $T$, meaning that all sub-level sets of $\EEE[T_i(\pi)]$ are convex. 

\begin{lemma}\label{lemma:quasi-convexity}
The expected delay at a station $\EEE[T_i(\pi)]$, given by~\eqref{eq:exp_break_cyclic}, is quasi-convex in $T$ for fixed $\{\lambda_i\}$ and $T_{tr}$. 
\end{lemma}
\noindent{\sc Proof.} See Appendix~\ref{appendix:proofs} for the mostly technical proof. ~\qed

Another important property of $\EEE[T_i(\pi)]$ is its monotonic dependency over $\lambda_i$, holding other parameters fixed. 
\begin{lemma}\label{lemma:monotonic_lambda}
For fixed $\sigma$, policy period $T$, and policy $\pi$ given by~\eqref{eq:sol1}, $\EEE[T_i(\pi)]$ increases monotonically as $\lambda_i$ increases. 
\end{lemma}
\noindent{\sc Proof.} Plugging $\tobs := T - \ttr$ and $\sigma := 1/(\sum_{i=1}^n (1/\lambda_i))$ into~\eqref{eq:exp_break_cyclic} and treating it as a function of $\lambda_i$ with $T, \ttr$, and $\sigma$ all fixed, we get
\begin{equation}\label{equation_lambda}
f_N(\lambda_i) = \displaystyle\frac{2}{\lambda_i} + \frac{T - \frac{\sigma \tobs}{\lambda_i}(1 + e^{-\sigma \tobs})}{1 - e^{-\sigma \tobs}},
\end{equation}
the derivative of which is
\begin{equation}\label{equation_lambda_derivative}
f_N'(\lambda_i) = \displaystyle\frac{\sigma \tobs e^{-\sigma \tobs} + \sigma \tobs + 2e^{-\sigma \tobs} - 2}{\lambda_i^2(1-e^{-\sigma \tobs})},
\end{equation}
which is strictly positive for all positive $\sigma \tobs$ and arbitrary positive $\lambda_i$, implying that $f_N(\lambda_i)$ increases monotonically with respect to $\lambda_i$. 
~\qed

\noindent{\sc Proof of Theorem~\ref{theorem:optimality}.} Lemma~\ref{lemma_continuum} and Lemma~\ref{l:exp_break_cyclic} show that the optimal policy period is given by
$$
\begin{array}{l}
T^* := \underset{T > \ttr }{\arg\min}\, \max_{i}\EEE[T_i(\pi)]\\
\quad = \underset{T > \ttr }{\arg\min}\, \max_{i}\Big[\displaystyle\displaystyle\frac{2}{\lambda_i} +  \frac{T - t_i - t_ie^{-\lambda_it_i}}{1 - e^{-\lambda_it_i}}\Big].
\end{array}
$$

By monotonicity of $\EEE[T_i(\pi)]$ with respect to $\lambda_i$ (Lemma~\ref{lemma:monotonic_lambda}), $\max_{i}\EEE[T_i(\pi)]$ is simply $\EEE[T_i(\pi)]$ for the station $i$ with the largest $\lambda_i$. This reduces computing $T^*$ to finding the minimum on a single function, which is a quasi-convex function by Lemma~\ref{lemma:quasi-convexity}.~\qed

\subsection{Uniqueness of Optimal Solution}
For a multi-objective optimization problem, results like Theorem~\ref{theorem:optimality} generally only give one optimal solution on the Pareto front with a potentially continuum of optimal solutions. However, the policy given by Theorem~\ref{theorem:optimality} is in fact the {\em unique} optimal solution, due to Theorem~\ref{theorem:uniqueness}. 

\begin{theorem}[Uniqueness of Optimal Solution]\label{theorem:uniqueness} The optimal policy $\pi^*$ provided by Theorem~\ref{theorem:optimality} is the unique policy that solves Problem~\ref{problem:cycle}. 
\end{theorem}
\noindent {\sc Proof}. We assume that we work with a fixed problem instance and assume the optimal policy computed by Algorithm~\ref{algorithm:schedule} has a period of $T^*$. Theorem~\ref{theorem:optimality} shows that $J_1 \le 1/n$ and can always reach $1/n$. To show that no other policy other than $\pi^*$ lies on the Pareto front, we need to show that no policy with fixed $J_1 < 1/n$ yields better value on $J_2$. 

Assume instead that there is another Pareto optimal solution $\pi' = (t_1', \ldots, t_n') \ne \pi^*$ for the same problem instance with $J_1(\pi') = c < 1/n$. Let the period of $\pi'$ be $T'$ and let $\tobs' = T' - \ttr$. Let $\pi''$ be the cyclic policy also with cycle period $T'$ such that $J_1(\pi'') = 1/n$. Note that $\pi''$ is unique and $\pi'' = \pi^*$ if $T' = T^*$. For $\pi'$ to be on the Pareto front, because $J_1(\pi') = c < 1/n = J_1(\pi^*)$, one must have $J_2(\pi') < J_2(\pi^*)$; we show that on the contrary we always have $J_2(\pi') > J_2(\pi'') \ge J_2(\pi^*)$, in which the last inequality is clear. We are left to show $J_2(\pi') > J_2(\pi'')$.

Since $J_1(\pi') = \min_i\alpha_i(\pi')$, we may assume $J_1(\pi') = \alpha_1(\pi') = c$. This implies that $\lambda_1t_1' \le \lambda_it_i'$ for all $2 \le i \le n$. To satisfy constraint~\eqref{equation:objective_one_constraint}, which translates to $\min_{i \ge 2} (\lambda_it_i' - \lambda_1t_1')$, we must have $\alpha_2(\pi') = \ldots = \alpha_{n}(\pi') > 1/n$ because having more $\alpha_i(\pi') < 1/n, i \ge 2$ will only increase $\min_{i \ge 2} (\lambda_it_i' - \lambda_1t_1')$ (note that $\tobs'$ is fixed). We then compute $\alpha_i(\pi') = (1 - c)/(n -1)$ for $i \ge 2$ and 
\begin{align}\label{equation:lt2}
\frac{\lambda_it_i'}{\lambda_1t_1'} = \frac{\alpha_i(\pi')}{\alpha_1(\pi')} =  \frac{1-c}{c(n-1)}.
\end{align}

Also we have $\lambda_2t_2' = \ldots = \lambda_nt_n'$ for $i \ge 2$. Solving this with the constraint $\sum_{i = 2}^n t_i' = \tobs' - t_1'$ gives us for $i \ge 2$,
\begin{align}\label{equation:lt1}
\lambda_it_i' = \frac{\tobs' - t_1'}{\sum_{j= 2}^n\frac{1}{\lambda_j}} = \frac{\tobs'}{\sum_{j= 2}^n\frac{1}{\lambda_j}} - \frac{1}{\lambda_1\sum_{j= 2}^n\frac{1}{\lambda_j}}\lambda_1t_1'.
\end{align}

Putting~\eqref{equation:lt2} and~\eqref{equation:lt1} together, we get
\begin{align}\label{equation:l1t1}
\lambda_1 t_1' = \frac{c(n-1)\tobs'}{(1-c)\sum_{i = 2}^n\frac{1}{\lambda_i} + c(n-1)\frac{1}{\lambda_1}}.
\end{align}

Also from~\eqref{equation:lt2}, 
\begin{align}\label{equation:deltadiff}
\Delta_{1i}(\pi') = \lambda_it_i' - \lambda_1t_1' = \frac{1-c}{c(n-1)}\lambda_1t_1' - \lambda_1t_1' = \frac{1-cn}{c(n-1)}\lambda_1 t_1'.
\end{align}

Plugging~\eqref{equation:l1t1} into~\eqref{equation:deltadiff} gives us that for some constant $C$, 
$$
(\Delta_{1i}(\pi'))^{-1} = C((1-c)(\frac{1}{\lambda_2} + \ldots + \frac{1}{\lambda_n}) + c(n-1)\frac{1}{\lambda_1}).
$$
Because $(1-c) > (n-1)/n > c(n-1)$, to minimize $\Delta_{1i}(\pi')$ or maximize its inverse, $\lambda_1$ must equal $\lmax$. This implies that $t_1' < t_1''$ ({\em i.e.}, the time spent per cycle at a station with $\lmax$ is less in $\pi'$ than in $\pi''$). Therefore, because~\eqref{eq:exp_break_cyclic} monotonically decreases as $t_i > 0$ increases when the policy period $T$ is fixed, $J_2(\pi') > J_2(\pi'')$. ~\qed

\section{Convergence and Robustness Properties of the Optimal Scheduling Policy}\label{section:property}

In this section, we prove two important ``goodness'' properties of the unique optimal cyclic policy for Problem~\ref{problem:cycle}, namely, the {\em convergence rate} of the policy toward its desired steady-state behavior and the {\em robustness} of the policy with respect to small perturbations of the input parameters. 

\subsection{Convergence Rate toward Stead-State Behavior}
Given an optimal policy $\pi^*$ for Problem~\ref{problem:cycle}, we have proved that the total number of events observed at a station $i$, divided by the number of all observed events, converges to $\alpha_i(\pi^*)$ in expectation ({\em i.e.}, given infinite amount of time). In practice, the execution of a monitoring policy must start at some point of time (instead of at $-\infty$) and only lasts a finite amount of time. Therefore, it is generally desirable that the sample averages converge quickly to their respective expected values. Here, we characterize the convergence rate of the fraction of observations with respect to the number of executed policy cycles. We do so by looking at the variance of these ratios around their expected values. 

\begin{theorem}[Convergence of the Fraction of Observations] \label{theorem:convergence}
Suppose the optimal policy for Problem~\ref{problem:cycle} is executed for $m$ cycles, that is, for $m\, T^*$ amount of time in which $T^*$ is the optimal policy's period. Then, the standard deviation of the fraction of total observations up until time $m\, T^*$ acquired at any particular station is 
$$
\frac{1}{\sqrt{m\sigma \, \tobs^*}},
$$
in which $\sigma$ is the harmonic sum of the arrival rates.\footnote{In computing this measure, we look at the fraction of the total number of observations in one station versus the total number of observations acquired up until time $m\,T^*$.}
\end{theorem}
\noindent{\sc Proof.} For convenience, assume that $t$ is an integer multiple of cycle time $T^*$, {\em i.e.}, $t = mT^*$, $m = 1, 2, \ldots$. For a fixed $m$, the Poisson process at station $i$ is equivalent to a Poisson {\em distribution} with arrival rate
$$
\lambda = k\lambda_i t_i = \frac{m(T^*-\ttr )}{\sum_j\frac{1}{\lambda_j}},
$$
which means that the variance of the number of data points observed is simply $\lambda$. The standard deviation of this Poisson distribution is then $\sqrt{\lambda}$, yielding a ratio of 
\begin{equation}\label{equation_convergence_rate}
\displaystyle\frac{\sqrt{\lambda}}{\lambda} = \sqrt{\frac{1}{\lambda}} = \sqrt{\frac{\sum_i\frac{1}{\lambda_i}}{m(T^*-\ttr )}} = \sqrt{\frac{1}{m\sigma \tobs }}. 
\end{equation}
~\qed

We note that the standard deviation given by~\eqref{equation_convergence_rate} is independent of the particular station. That is, the optimal schedule is such that the convergence occurs at the same rate across all stations. The theorem states that this standard deviation is inversely proportional to the square root of the number of cycles the schedule is executed, which is fairly reasonable.

\subsection{Robustness of Optimal Policy}
Another important issue related to solution soundness is its robustness. Under the particular context of this paper, it is desirable to ensure that the computed policy is robust with respect to small perturbations in the input parameters. Here, input parameters to our problem are $\tau_{i,j}$, $\{\lambda_i\}$, and an ordering of the stations. Since the ordering is a combinatorial object, it does not directly subject to perturbations. Therefore, we focus on the other two sets of parameters, which are continuous variables and can be readily perturbed. 

\subsubsection*{Robustness with respect to perturbations in $\{\lambda_i\}$} We show that, when the optimal policy is deployed, the change in the expected delay $\EEE[T_i(\pi)]$ at a station $i$ is bounded with respect to small changes in $\lmax$. Furthermore, the rate of the change is fairly limited at nearly all stations.  

\begin{theorem}[Robustness w.r.t Arrival Rate] \label{theorem:robustness}
Let us denote the delay at station $i$ under the optimal schedule as a function of $\lambda_i$ by letting $f_N(\lambda_i) := \EEE[T_i (\pi)]$. Holding $\sigma$ fixed and letting $x := \sigma\tobs$, then
$$
\big(\frac{\Delta(f_N(\lambda_i))}{f_N(\lambda_i)}\big)\big/\big(\frac{\Delta \lambda_i}{\lambda_i}\big) < - \frac{2 - 2 e^{-x} - x(1+e^{-x})}{2-2e^{-x} - x(1+e^{-x})+\frac{\lambda_i}{\sigma}x},
$$
the RHS of which is always upper bounded, and takes values in $(0,1)$ for all $x \in (0,\infty)$ and $\lambda_i \ge \lmin := \min_j \lambda_j$. 
\end{theorem}
\noindent{\sc Proof.} Quantitatively, we want to show that $f_N(\lambda_i)$ (see~\eqref{equation_lambda}) does not change fast as $\lambda_i$ varies. More formally, we seek to prove that $\Delta(f_N(\lambda_i))/{f_N(\lambda_i)}$ is small for small $\Delta\lambda_i/\lambda_i$. Through Taylor expansion, 
$$
\frac{\Delta(f_N(\lambda_i))}{f_N(\lambda_i)} \approx \frac{f_N'(\lambda_i)\Delta \lambda_i}{f_N(\lambda_i)} = \frac{\lambda_if_N'(\lambda_i)}{f_N(\lambda_i)}\frac{\Delta\lambda_i}{\lambda_i}.
$$
By~\eqref{equation_lambda} and \eqref{equation_lambda_derivative}, 
$$
\begin{array}{l}
\displaystyle\frac{\lambda_if_N'(\lambda_i)}{f_N(\lambda_i)} = -\frac{2-2e^{-\sigma \tobs} - \sigma \tobs(1+e^{-\sigma \tobs})}{2-2e^{-\sigma \tobs} - \sigma \tobs(1+e^{-\sigma \tobs}) + \lambda_iT} \\
\displaystyle\quad \overset{x := \sigma \tobs}{=} -\frac{2-2e^{-x} - x(1+e^{-x})}{2-2e^{-x} - x(1+e^{-x}) + \frac{x\lambda_i}{\sigma} + \lambda_i\ttr },
\end{array}
$$
in which $x > 0$ and $\ttr  > 0$. Because $\lambda_i/\sigma > 1$, $\lambda_if_N'(\lambda_i)/f_N(\lambda_i)$ can be shown to be upper bounded by $1/(\lambda_i/\sigma - 1)$. In particular, for all $\lambda_i > \lmin$, $\lambda_i/\sigma > 2$ holds, yielding
$$
\begin{array}{l}
\displaystyle\frac{\lambda_i f_N'(\lambda_i)}{f_N(\lambda_i)} < -\frac{2-2e^{-x} - x(1+e^{-x})}{2-2e^{-x} - x(1+e^{-x}) + 2x},
\end{array}
$$
which takes value in $(0, 1)$ for all $x \in (0, \infty)$. If $\lmax = \lambda_1 = \ldots = \lambda_n = \lmin$, then we can similarly show that $\lambda_if_N'(\lambda_i)/f_N(\lambda_i) < 1$.  ~\qed

Since small relative changes to $\lambda_{max}$ only induce relative changes of smaller or equal magnitude to the corresponding expected delay by Theorem~\ref{theorem:robustness}, the optimal policy is robust with respect to perturbations to event arrival rates. 

\subsubsection*{Robustness with respect to perturbations in $\{\tau_{i,j}\}$} Now suppose instead that elements of $\{\tau_{i,j}\}$ are perturbed. The only relevant change induced by these perturbations is a perturbation to $\ttr$, the total travel time in a policy period. Perturbing $\ttr$ causes a change in $\tobs^*$, which is determined by the largest arrival rate $\lmax$. We characterize the relative magnitude of this effect in the theorem below.
\begin{theorem}[Robustness w.r.t. Travel Time] \label{theorem:robustness-ttr}
Let us denote the delay at the station with the maximum arrival rate as a function of $\ttr$ by letting 
\begin{align}\label{equation:ft}
f_{\tobs}(\ttr) := \frac{2}{\lmax} + \frac{(\tobs + \ttr) \lmax - \sigma \tobs (1 + e^{-\sigma \, \tobs}) }{(1 - e^{- \sigma \tobs})\lmax},
\end{align}
Holding $\{\lambda_i\}$ and $\tobs$ fixed, then
$$
\big(\frac{\Delta(f_{\tobs}(\ttr))}{f_{\tobs}(\ttr)}\big)\big/\big(\frac{\Delta \ttr}{\ttr}\big) \in (0, 1).
$$
\end{theorem}
\noindent{\sc Proof.} Following the proof of Theorem~\ref{theorem:robustness} and letting $x:=\sigma\tobs$, we compute 
$$
\begin{array}{l}
\displaystyle\big(\frac{\Delta(f_{\tobs}(\ttr))}{f_{\tobs}(\ttr)}\big)\big/\big(\frac{\Delta \ttr}{\ttr}\big) \approx \frac{\ttr f_{\tobs}'(\ttr )}{f_{\tobs}(\ttr)} \\
\displaystyle\quad = \frac{\lmax\ttr(1-e^{-x})}{2(1-e^{-x}) - x(1 + e^{-x}) + \lmax\tobs + \lmax\ttr}\\
\displaystyle\quad < \frac{\lmax\ttr(1-e^{-x})}{2(1-e^{-x}) - x(1 + e^{-x}) + x + \lmax\ttr}\\
\displaystyle\quad = \frac{\lmax\ttr(1-e^{-x})}{2 - 2e^{-x} - xe^{-x} + \lmax\ttr}.
\end{array}
$$
The inequality is due to $\lmax > \sigma$. Because $2 - 2e^{-x} - xe^{-x} > 0$ for all $x = \sigma\tobs > 0$ and $0 < 1-e^{-x} < 1$, we conclude that $\ttr f_{\tobs}'(\ttr )/f_{\tobs}(\ttr) \in (0, 1)$. ~\qed

With Theorem~\ref{theorem:robustness-ttr}, we conclude that the optimal policy is robust with respect to perturbing the travel times, $\{\tau_{i,j}\}$. 

\textbf{Remark.} We note that the results from Sections~\ref{section:theory} and~\ref{section:property} continue to hold when the visiting order of the stations is not predetermined, due to the fact that travel times only matter as a whole ({\em i.e.}, through $\ttr$). The only non-essential difference is that they may be multiple optimal policies yielding the same $J_1$ and $J_2$ values, because there may be multiple TSP tours for a given problem instance. Such {\em degenerate} cases are however very rare.\footnote{It is possible to show that such cases have zero measure with mild assumption on the station distribution.} Our analysis also implies that Problem~\ref{problem:general} is NP-hard because it contains TSP. 

\section{Computing the Optimal Scheduling Policy: Algorithm and Complexity Analysis}\label{section:algorithm}
In this section, we first provide an algorithm for solving Problem~\ref{problem:cycle} and characterize its performance. Then, building on this algorithm and robustness results from Section~\ref{section:property}, we provide a polynomial-time approximation scheme (PTAS) for solving Problem~\ref{problem:general}. 

\subsection{Algorithm for Computing Cyclic Policy with Predetermined Station Visiting Order}
The pseudo-code for computing the unique cyclic policy $\pi^*$ solving Problem~\ref{problem:cycle} is given in Algorithm~\ref{algorithm:schedule}, as a direct consequence of Theorem~\ref{theorem:optimality} and Theorem~\ref{theorem:uniqueness}. First, in Lines~\ref{line:schedule:statistics_start}-\ref{line:schedule:statistics_end}, the algorithm computes two useful statistics, namely, the maximum arrival rate (denoted by $\lmax$) and $\sigma$. Then, in Line~\ref{line:schedule:quasiconvex}, the algorithm proceeds by solving an optimization problem in one variable, $\tobs$. At this step, the algorithm computes the optimal total observation time denoted by $\tobs^*$. Finally, the algorithm computes the optimal total cycle period $T^* = T^*_\mathrm{obs} + \ttr$ in Line~\ref{line:schedule:cycle_time} and the optimal observation time for the individual stations in Lines~\ref{line:algorithm:optimal_observation_times_start}-\ref{line:algorithm:optimal_observation_times_end}.

\def\algoOrder{{\sc CompOptOrdered}}
\def\algoNoOrder{{\sc CompOptUnOrdered}}
\begin{algorithm}
    \SetKwInOut{Input}{Input}
    \SetKwInOut{Output}{Output}
    \SetKwComment{Comment}{\%}{}
    \Input{$(\lambda_1,\ldots,\lambda_n)$: ordered arrival rates \\ 
		$\{\tau_{i,j}\}$: the travel times }
    \Output{$\pi^* = (t_1^*,t_2^*,\dots, t_n^*)$: the optimal policy}

\vspace{0.15in}
		
\Comment{\small Compute relevant statistics}

\vspace{0.025in}

$\lmax \leftarrow \max\limits_{1\le i \le n} \lambda_i$ \label{line:schedule:statistics_start}\Comment*{\small The maximum of $\lambda_i$'s}
$\sigma \leftarrow \left(\sum\nolimits_{i =1}^n \lambda_i^{-1}\right)^{-1}$ \label{line:schedule:statistics_end}\Comment*{\small The harmonic sum of $\lambda_i$'s}

\vspace{0.15in}

\Comment{\small Solve a quasi-convex optimization problem}
$\tobs^* \leftarrow \underset{\tobs > 0}{\arg\min} \left(\frac{2}{\lmax} + \frac{(\tobs + \ttr) \lmax - \sigma \tobs (1 + e^{-\sigma \, \tobs}) }{(1 - e^{- \sigma \tobs})\lmax} \right)$\;  \label{line:schedule:quasiconvex}

\vspace{0.15in}

\Comment{\small Calculate the optimal policy}
$T^* \leftarrow \tobs^* + \ttr$  \label{line:schedule:cycle_time}\hfill\Comment{\small Calculate optimal cycle time}
\For{$i \in \{1,2,\dots, n\}$ }{ \label{line:algorithm:optimal_observation_times_start}
	\Comment{\small Calculate optimal observation times.}
	$t_i^* \leftarrow \frac{\sigma}{\lambda_i} \tobs^*$  \label{line:algorithm:optimal_observation_times_end}
}
\Return{$\pi^* = (t_1^*,t_2^*, \dots, t_n^*)$}
\caption{\algoOrder} \label{algorithm:schedule}
\end{algorithm}

We emphasize that the optimization problem in Line~\ref{line:schedule:quasiconvex} of Algorithm~\ref{algorithm:schedule} is a quasi-convex optimization problem in one variable by Theorem~\ref{theorem:optimality}, which can be solved efficiently in multiple ways. For example, because the value of the function to be minimized can be computed analytically, we may apply the bisection method or the Newton-Raphson method to compute $\pi^*$ very efficiently. 

On the side of computational complexities of Algorithm~\ref{algorithm:schedule}, the following theorem is immediate. We measure the computational complexity of the algorithm by the number of steps executed by the algorithm. A single step is either a comparison, an addition, or a multiplication operation. 
\begin{theorem}[Complexity of Computing Optimal Schedule] \label{theorem:computational_complexity} The number of steps performed by Algorithm~\ref{algorithm:schedule} is $O(n)$, in which $n$ is the number of stations. Moreover, if $\lmax = \max_i\lambda_i$ and the harmonic sum $\sigma = 1/\sum_{i=1}^n(1/\lambda_i)$ are known, the optimal cycle time can be computed in constant time. 
\end{theorem}

Now, we consider {\em online} problem instances, in which new stations are added or other existing ones are removed, on the fly, from the list of stations to be serviced. The task is to construct the optimal schedule and maintain it as the list of stations to be serviced changes. 

First consider the problem with addition only. In that case, the online algorithm can be described as follows. At any given time, the algorithm maintains the maximum rate $\lmax$ and the harmonic sum $\sigma = (\sum_{i=1}^n \lambda_i^{-1})^{-1}$. Let $\lambda_\mathrm{new}$ denote the event arrival rate for the new station. Then, the new statistics, denoted by $\lmax'$ and $\sigma'$, are computed as follows: 
\begin{align*}
\lmax' &\leftarrow \max\{\lmax, \lambda_\mathrm{new}\}\\
\sigma' &\leftarrow \left(\,\sigma^{-1} + 1/\lambda_\mathrm{new}\right)^{-1}
\end{align*}
Then, solve the quasi-convex optimization problem in Line~\ref{line:schedule:quasiconvex} of Algorithm~\ref{algorithm:schedule} to compute the optimal cycle time. Notice that these computations (the update and the solution of the quasi-convex optimization problem) can be executed in constant time. The running time of the algorithm that updates the optimal schedule time is independent of the number of stations. 
Second, consider the case when a new station may be added or an existing one can be removed. In this case, clearly the statistic $\sigma$ can still be updated in constant time. However, maintaining the statistic $\lmax$ is harder in the case of removals, since removing the station with rate $\lmax$ requires looking through the remaining stations to find the station with the largest event arrival rate. This can not be done in constant time. Yet, an ordered list of the stations can be maintained in logarithmic time. More precisely, the robot maintains an ordered list of stations such that the ordering is with respect to the event arrival rates $\lambda_i$. Adding a new station or removing a station from this can be performed in $\log (n)$ time, in which $n$ is the number of stations. Once addition or removal is performed, the maximum event arrival rate, $\lmax$, can be updated immediately. Hence, the overall update algorithm requires logarithmic time in the number of stations.

We summarize this as a corollary of our previous results. 

\begin{corollary}[Online Complexity]
Consider the case in which new stations are added to the list of stations to be served, on the fly. When a new station is added to a list of stations to be observed, the optimal scheduling policy can be updated in constant time, independent of the number of existing stations. Consider the case when the stations are both added to and removed from a list of $n$ stations to be served. Then, when a new station is added or removed, the optimal scheduling policy can be updated in $O(\log(n))$ time.
\end{corollary}

{\bf Remark}. First, the space complexity, {\em i.e.}, the amount of memory required to maintain the optimal cycle time, is constant when there are only additions. The space complexity is linear when there are removals as well. Second, clearly, solely updating the cycle time is not enough for implementing the optimal schedule; one must also update the time spent in each station. However, from a practical point of view, the time spent in each station can be updated as the robot travels to these destinations. This strategy should work well as long as the robot has computational power to evaluate Line~\ref{line:algorithm:optimal_observation_times_end} of Algorithm~\ref{algorithm:schedule} (which requires two multiplications and one addition) during the time it spends at station $i-1$ and the time it travels to station $i$. In other words, the robot can compute the optimal cycle time $T^*$ and start its monitoring of the stations. Right after $T^*$ is computed, the robot can start the implementation of the plan. It computes $t_1^*$ on the way to station 1 and when waiting at station 1, and so on.

\subsection{Computing Optimal Cyclic Policies without a Predetermined Station Visiting Order}
Our algorithm for computing a $(1+\epsilon)$-optimal solution for Problem~\ref{problem:general}, outlined in Algorithm~\ref{algorithm:schedule-no-order}, is a simple routine sequentially calling a TSP subroutine and then Algorithm~\ref{algorithm:schedule}. The flow of Algorithm~\ref{algorithm:schedule-no-order} is straightforward to understand. The challenge is to show that a $(1+\epsilon)$-optimal TSP solution is all we need for computing a $(1+\epsilon)$-optimal solution to Problem~\ref{problem:general}. We now prove the correctness and the stated time complexity of Algorithm~\ref{algorithm:schedule-no-order}.

\begin{algorithm}
    \SetKwInOut{Input}{Input}
    \SetKwInOut{Output}{Output}
    \SetKwComment{Comment}{\%}{}
    \Input{$\{\lambda_1,\ldots,\lambda_n\}$: the arrival rates, unordered \\ 
		$\{\tau_{i,j}\}$: the travel times }
    \Output{$\pi^* = ((k_1^*, t_1^*),\dots, (k_n^*, t_n^*))$: the optimal policy}

\vspace{0.15in}
		
\Comment{\small Compute an approximate TSP route}

\vspace{0.025in}

Using $\{\tau_{i,j}\}$, compute a $(1 + \epsilon)$-optimal TSP solution over the distances, yielding $(k_1^*, \ldots, k_n^*)$\label{a2l1}

\vspace{0.1in}

\Comment{\small Call the algorithm for Problem~\ref{problem:cycle}}
$(\lambda_1, \ldots, \lambda_n \leftarrow (\lambda_{k_1^*}, \ldots, \lambda_{k_n^*})$\Comment*{\small Reorder $\lambda_i$'s}\label{a2l2}
$(t_1^*, \ldots, t_n^*) \leftarrow$ \algoOrder$((\lambda_1, \ldots, \lambda_n), \{\tau_{i,j}\})$\label{a2l3}

\vspace{0.1in}
\Return{$\pi^* = ((k_1^*, t_1^*),\dots, (k_n^*, t_n^*))$}
\caption{\algoNoOrder} \label{algorithm:schedule-no-order}
\end{algorithm}

\begin{theorem}[PTAS for Unordered Stations]\label{theorem:ptas}
Fixing a real number $\epsilon > 0$, a $(1 + \epsilon)$-optimal policy can be computed for Problem~\ref{problem:general} in time polynomial in $n$, the number of stations. 
\end{theorem}
\noindent{\sc Proof.} Suppose that the optimal total travel time is $\ttr^*$ and Line~\ref{a2l1} of Algorithm~\ref{algorithm:schedule-no-order} computes a solution with total travel time $\ttr' = (1+\epsilon)\ttr^*$. The optimal policies computed using  $\ttr^*$ and $\ttr'$ (and the associated visiting order), computed by Algorithm~\ref{algorithm:schedule}, have the same optimal $J_1$ value. Let the optimal values of $J_2$ for these two polices be $J_2^*$ and $J_2'$, respectively. We further let the optimal total observation times per cycle corresponding to $\ttr^*$ and $\ttr'$ be $\tobs^*$ and $\tobs'$, respectively. For convenience, we reuse the definition of $f_{\tobs}(\ttr)$ given by~\eqref{equation:ft}, which gives us
$$
f_{\tobs^*}(\ttr) = \frac{2}{\lmax} + \frac{(\tobs^* + \ttr) \lmax - \sigma \tobs^* (1 + e^{-\sigma \, \tobs^*}) }{(1 - e^{- \sigma \tobs^*})\lmax}.
$$

We also know that $J_2^* = f_{\tobs^*}(\ttr^*)$ by definition. By Theorem~\ref{theorem:robustness-ttr}, in particular that $\ttr f_{\tobs}'(\ttr )/f_{\tobs}(\ttr) \in (0, 1)$, we have
$$
\begin{array}{l}
\displaystyle f_{\tobs^*}(\ttr') = f_{\tobs^*}((1+\epsilon)\ttr^*) \approx \displaystyle f_{\tobs^*}(\ttr^*) + \epsilon T^* f_{\tobs^*}'(\ttr^*) \\
\displaystyle \quad < f_{\tobs^*}(\ttr^*) + \epsilon f_{\tobs^*}(\ttr^*)  = (1 + \epsilon)f_{\tobs^*}(\ttr^*) \\
\end{array}
$$

On the other hand, it is straightforward to see that $f_{\tobs}(\ttr)$ is monotonically increasing in $\ttr$ since 
$$
f_{\tobs}'(\ttr) = \frac{1}{1 - e^{- \sigma \tobs}} > 0.
$$

Therefore, for all $\tobs > 0$, we have
$$
f_{\tobs}(\ttr') \ge f_{\tobs}(\ttr^*) \ge f_{\tobs^*}(\ttr^*) = J_2^*
$$
and
$$
J_2' = \min_{T > \ttr'} f_{\tobs}(\ttr') \le f_{\tobs^*}(\ttr') < (1 + \epsilon)f_{\tobs^*}(\ttr^*) = (1 + \epsilon)J_2^*.
$$

We conclude that $J_2^* \le J_2' \le (1+\epsilon) J_2^*$. That is, Algorithm~\ref{algorithm:schedule-no-order} produces a $(1+\epsilon)$-optimal solution to Problem~\ref{problem:general}. To achieve the desired polynomial-time complexity, because Algorithm~\ref{algorithm:schedule} takes linear time, we only need a polynomial-time approximation algorithm for computing a $(1 + \epsilon)$-optimal solution to the embedded TSP problem. Such a PTAS is provided in \cite{Aro98}. 
~\qed

\textbf{Remark.} It is often the case that PTAS does not yield practical polynomial-time algorithm. Luckily, this does not present an issue for us. Many fast TSP solvers exist. For example, LKH (Lin-Kernighan Heuristics) ~\cite{Hel00} can compute near optimal solutions for very larger TSP instances very quickly, even for more difficult TSP instances than Euclidean TSP, such as the asymmetric traveling salesmen problem (ATSP). Typical instances with thousands of cities can be solved in a few minutes to an accuracy of $1\%$ within the true optimal distance on a laptop.

\section{Computational Experiments}\label{section:experiment}
\begin{figure}[htp]
\begin{center}
\begin{tabular}{cc}
    \includegraphics[width=1.6in]{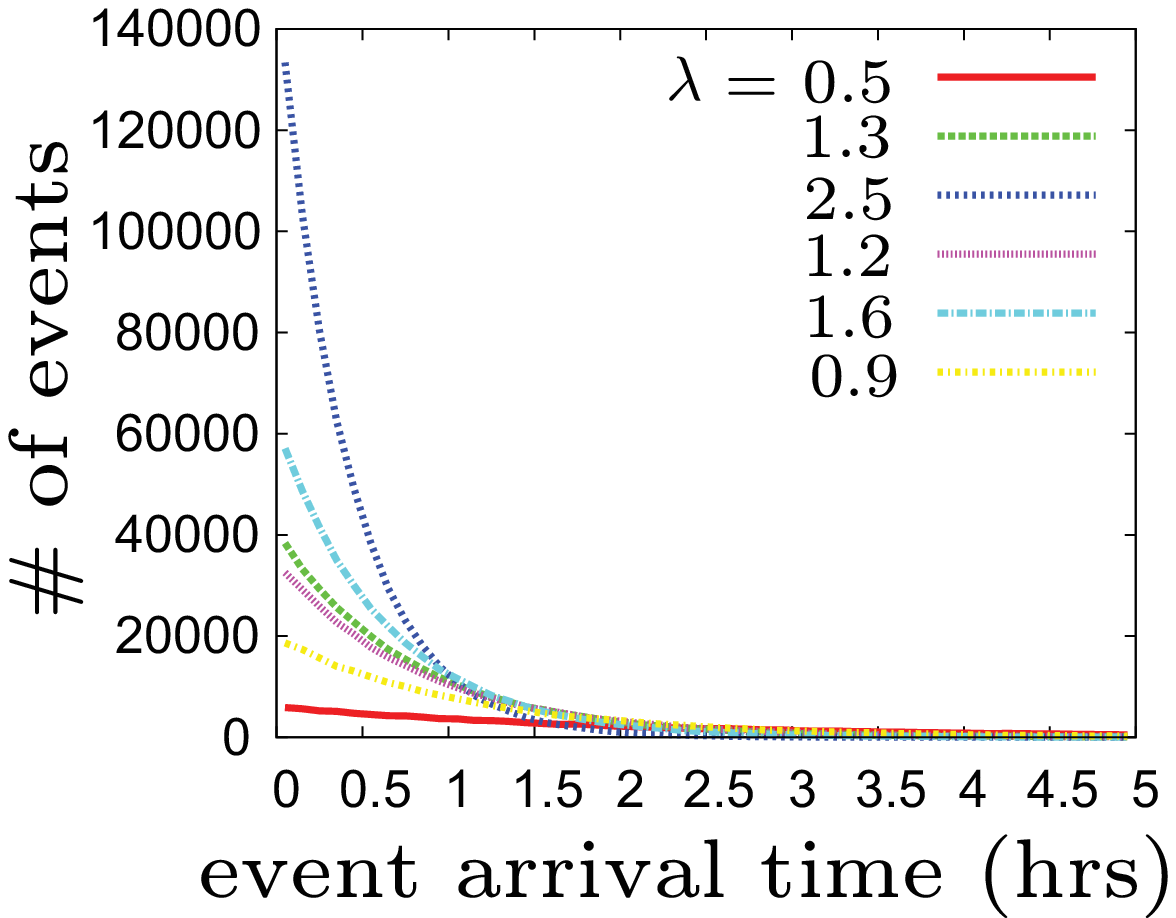}& 
    \includegraphics[width=1.6in]{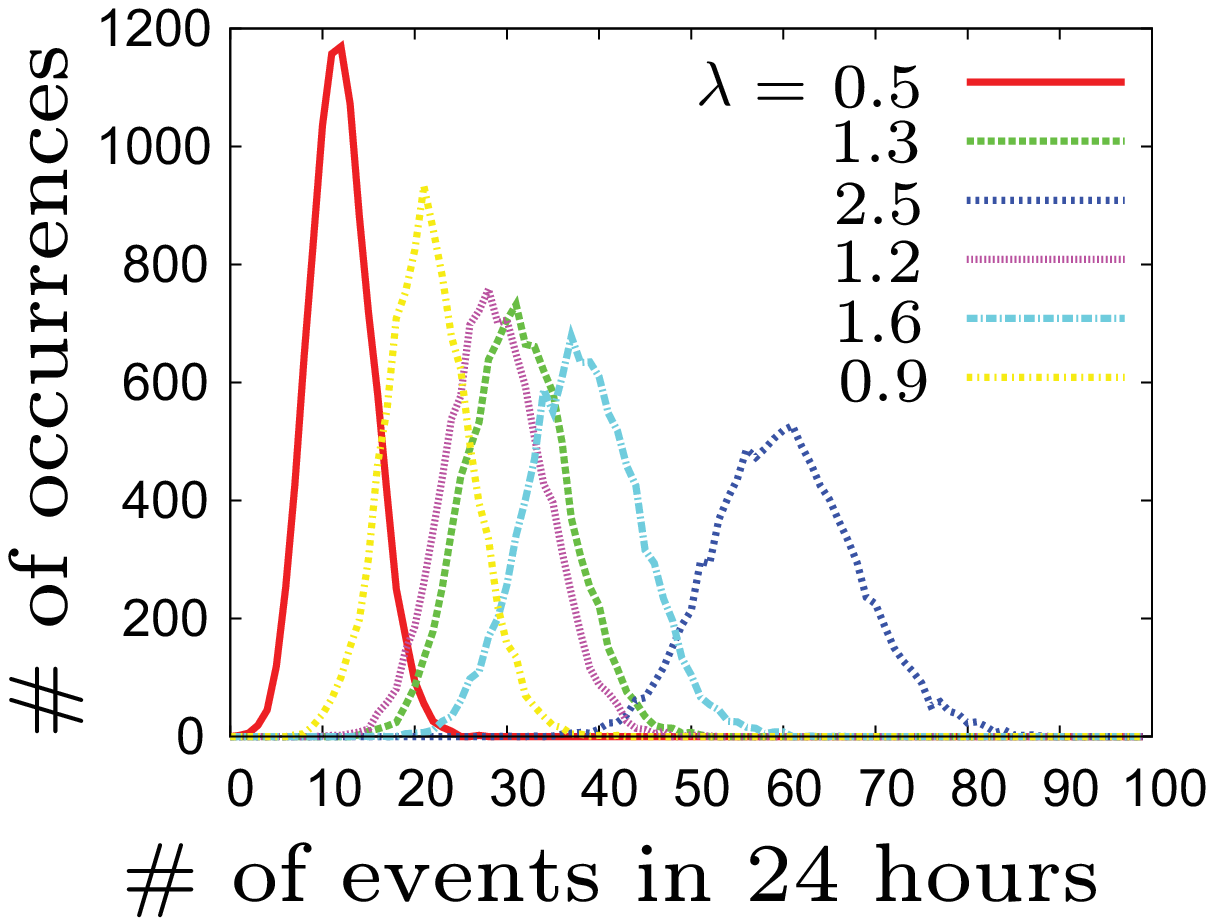} \\
    (a) & (b) 
\end{tabular}
\end{center}
\caption{\label{fig:params} (a) Histogram over the event arrival times since the last event arrival for the Poisson processes in our experiment over a time horizon of 10000 days. The bucket size (on the $x$ axis) is 0.1 hour.  (b) Histogram over the number of events arriving in an 24-hour window for the different Poisson processes over 10000 runs.}
\end{figure}
Recall the UAV monitoring application illustrated in Figure~\ref{fig:example}. The UAV is tasked with persistently monitoring six locations of interest and hover over each location for certain periods of time to capture events occuring at these locations. The input consists of the arrival rates for events at each station (denoted by $\lambda_i$) and the time needed for traveling between the stations (denoted by $\tau_{i,j}$). Table~\ref{table_param} lists these parameters. The time unit is hours (hr). Figure~\ref{fig:params} illustrates the stochastic nature of the event arrival times. Note that, in addition to the large range of average arrival rates at different stations ({\em e.g.}, events arrive at station 3 five times more frequent than they do at station 1), the stochastic arrival times can vary greatly within the same station. The UAV must balance the amount of data collected at all stations despite the different arrival rates while not incurring large delays in event observations between consecutive visits to the same location.

\begin{table}[htp]
\begin{center}
	\caption{\label{table_param} The ground truth (event arrival rates and travel times) used in our simulations. }
	\begin{tabular}{ccccccc}
   \hline\hline
	  & \multicolumn{6}{c}{Station} \\
	 \cline{2-7}
	   & 1 & 2 & 3 & 4 & 5 & 6\\
	 \hline
	 $\lambda_i$ (1/hr) & 0.5 & 1.3 & 2.5 & 1.2 & 1.6 & 0.9 \\
	 \hline
	 $\tau_{i, i+1 \mod 6}$ (hrs) & 0.15 & 0.25 & 0.1 & 0.3 & 0.2 & 0.2 \\
	 \hline\hline
	 \end{tabular}
\end{center}
\end{table}

The objective of the computational experiments is to confirm our theoretical findings given in Sections~\ref{section:theory} and~\ref{section:property}, and offer insights into the structure induced by the optimization problem. Note that we do not lose generality by focusing on the case with known station cyclic order, which we do here. First, we verify the optimality of the computed schedule (Theorem~\ref{theorem:optimality}). We show that the schedule returned by the algorithm indeed minimizes the delay across all stations in a balanced way in a practical example scenario. 
Second, we focus on the convergence properties (Theorem~\ref{theorem:convergence}). We show that, in the same scenario, the fraction of observations at each station converges to zero at the rate given in Theorem~\ref{theorem:convergence} as the execution time increases. 
Third, we look at the robustness of the optimal policy (Theorem~\ref{theorem:robustness}). We show that, in a variety of selected scenarios, the optimal policy is also robust with respect to the changes in event arrival statistics. We omit the simulation study on Theorem~\ref{theorem:robustness-ttr}, which yields robustness results very similar to that of Theorem~\ref{theorem:robustness}. 

We mention that the source code for our simulation software was developed using the Java programming language, and the simulation software itself was executed on a computer with a 1.3GHz Intel Core i5 CPU and 4GB memory. Mathematica 9 was used for computing the optimal policy using the gradient descent optimization procedure. As suggested by Theorem~\ref{theorem:computational_complexity}, on this computational hardware, the computation of the optimal policy is almost instantaneous, on the order of a few milliseconds.

\subsection{Computing the Optimal policy}
\begin{figure}[htp]
\begin{center}
    \includegraphics[width=2.8in]{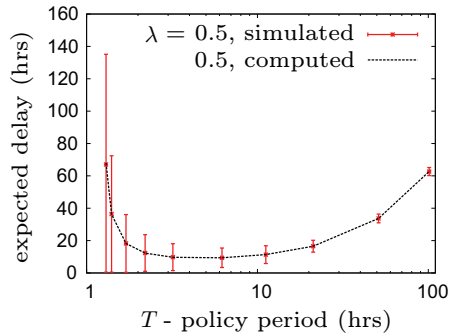} 
\end{center}
\caption{\label{fig:exp_delay} The simulated versus computed values for $\EEE[T_i(\pi)]$. We observe that the mean of the simulated runs agrees with the value computed directly from~\eqref{eq:exp_break_cyclic} for all choices of $T$'s, whereas the variance grows larger as $T \to \ttr$.}
\end{figure}

In this subsection, we focus on the optimality of the proposed schedule. First we show in simulations that our analysis correctly predicts the expected delay. Second, we compare the optimal schedule with an intuitive, but suboptimal policy.  

Below, we empirically check the correctness of Theorem~\ref{theorem:optimality} through simulations. Our first computational experiment validates~\eqref{eq:exp_break_cyclic} by performing both simulation and direct computation side by side and comparing the results, for the aforementioned case. In simulation, for each fixed $T \in \{1.3, 1.4, 1.7, 2.2, 3.2, 6.2, 11.2, 21.2, 51.2, 101.2\}$, we simulated the Poisson process for enough number of periods (roughly $2 \times 10^5$ in the worst case) to gather at $2000$ delays by simulating the policy. This gave us $2000$ samples of the random variable $T_i(\pi)$ from which we computed the mean and standard deviation. Direct computation based on~\eqref{eq:exp_break_cyclic} were also carried out. To avoid cluttering the presentation, only $\lambda = 0.5$ was used (plots for other $\lambda$ are similar). 
\begin{figure}[htp]
\begin{center}
    \includegraphics[width=2.8in]{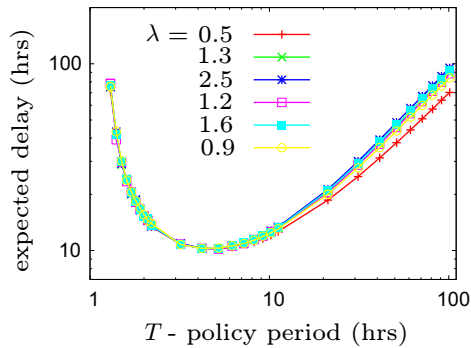} 
\end{center}
\caption{\label{fig:exp_delay_more} The computed $\EEE[T_i(\pi)]$ for $\lambda_1, \ldots, \lambda_6$ and $T \in [1.3, 101.2]$.}
\end{figure}
\begin{figure}[htp]
\begin{center}
    \includegraphics[width=2.8in]{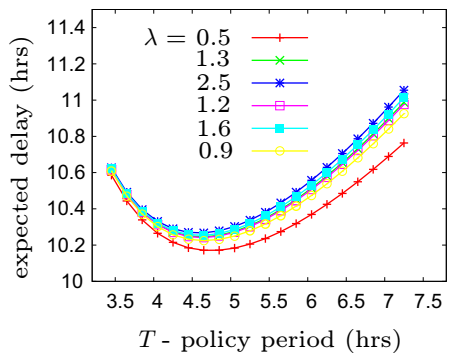} 
\end{center}
\caption{\label{fig:exp_delay_more_zoom} The computed $\EEE[T_i(\pi)]$ for $\lambda_1, \ldots, \lambda_6$ and $T \in [6.2, 10.2]$ with $\Delta T = 0.025$ increments.}
\end{figure}

The results of this simulation study are presented in Figure~\ref{fig:exp_delay}, in comparison with the optimal policy that is directly computed using the gradient descent procedure. Notice that the expected delay in simulation results match that of the computed policy exactly for all choices of $T$'s. We also observe from the simulation study that the variance of the delay increases as $T$ approaches $\ttr$. This should be intuitively clear, since, as $T - \ttr  \to 0^+$, the length of each observation window decreases when compared to $\ttr$; in fact, the ratio of the two approaches zero, which leads to the unbounded increase in the variance of the number of events observed in a given observation window. 

After empirically verifying that~\eqref{eq:exp_break_cyclic} is accurate, we shift our attention to the quasi-convexity of~\eqref{eq:exp_break_cyclic} and its monotonicity in $\lambda_i$. We compute $\EEE[T_i(\pi)]$ for all six $\lambda_i$'s and plot the result at two different scales in Figure \ref{fig:exp_delay_more} and \ref{fig:exp_delay_more_zoom}. Figure \ref{fig:exp_delay_more} shows that $\EEE[T_i(\pi)]$ is quasi-convex (in this case, convex) for all $\lambda_i$'s. Figure~\ref{fig:exp_delay_more_zoom}, the zoomed-in version of Figure \ref{fig:exp_delay_more}, further reveals that $\EEE[T_i(\pi)]$ depends on $\lambda_i$ monotonically for fixed period $T$, confirming the claim of Lemma~\ref{lemma:monotonic_lambda}. 

To compute the optimal cyclic patrolling policy's parameters, by Lemma~\ref{lemma:monotonic_lambda} we only need to look at $\EEE[T_i(\pi)]$ for $\lambda_i = 2.5$. The period $T$ that minimizes~\eqref{eq:exp_break_cyclic} for $\lambda_i = 2.5$ can be easily computed using standard gradient descent methods. Our computation yields $T^* = 4.59$. The corresponding policy is then defined by $\pi = $ $(1.18, 0.45, 0.24, 0.49, 0.37, 0.67)$. 

{\bf Remark}. We note that $\EEE[T_i(\pi)]$ is not always {\em convex}, contrary to what may be suggested by computational experiments ({\em e.g.}, Fig.~\ref{fig:exp_delay_more}). To see that $\EEE[T_i(\pi)]$ is quasi-convex, pick $n = 2$ as the number of stations with $\lambda_1 = 1$, $\lambda_2 = 4$, and $t_{12} = t_{21} = 0.0001$. For $T = 1.2502, 2.5002$, and $3.7502$, the optimal policies balancing the observed data and the corresponding $\EEE[T_1(\pi)]$ are given in Table~\ref{table_non_convexity}, form which one can easily verify that the point $(2.5002, 2.265)$ lies above the line connecting points $(1.2502, 1.814)$ and $(3.7502, 2.632)$, implying that $\EEE[T_1(\pi)]$ is non-convex on the interval $[1.2502, 3.7502]$. 
\begin{table}[htp]
\begin{center}
	\caption{\label{table_non_convexity} The expected delay for three policies for the same environments in the policies that optimize $J_1$}
	\begin{tabular}{cccc}
   \hline\hline
	 {No.} & $T$ & $\pi$ & $\EEE[T_1(\pi)]$ \\
	 \hline
	 1 & 1.2502 & $(1,0.25)$ & 1.814 \\
	 \hline
	 2 & 2.5002 & $(2,0.5)$ & 2.265\\
	 \hline
	 3 & 3.7502 & $(3,0.75)$ & 2.632\\
	 \hline\hline
	 \end{tabular}
\end{center}
\end{table}

\subsection{Performance on Non-Poisson Distributed Data}\label{subsection:non-poisson}
\begin{figure}[htp]
\begin{center}
    \includegraphics[width=2.8in]{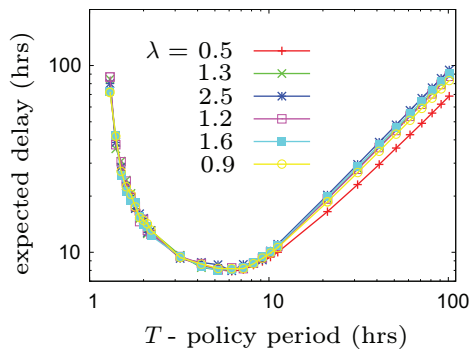} 
\end{center}
\caption{\label{figure:uniform} Simulated average delay using data generated over a variation of the uniform distribution.}
\end{figure}

As it is often the case that stochastic event arrival diverge from Poisson process, we are curious how our computed policy would perform on more realistic data. Because a large number of data points are needed to compute entities like average delay, instead of using real world data, we generated data to simulate processes like bus arrivals in the following manner. Using $\{\lambda_i\}$ from our main example, for each station $i$, we partition the time line into segments of length $1/\lambda_i$. In each segment, a uniformly random point is selected as the arrival time of an event at station $i$. Clearly, over this data set, the value of $J_1$ remains the same. When we use the policies that maximize $J_1$ to simulate $\EEE[T_i(\pi)]$ over this data set, we obtain results that are subsequently plotted in Figure~\ref{figure:uniform}. 

\begin{figure}[htp]
\begin{center}
    \includegraphics[width=2.8in]{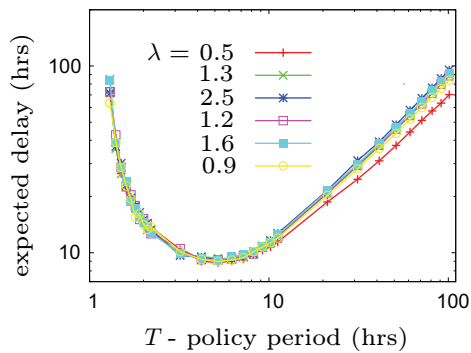} 
\end{center}
\caption{\label{figure:uniform-burst} Simulated average delay using data generated over a variation of the uniform distribution with bursts of arrivals.}
\end{figure}

We observe that the average delay is actually shorter in this case, implying a better optimal value for $J_2$. We also note that the general structure of $\EEE[T_i(\pi)]$ appears to remain the same, {\em i.e.}, largely convex. Then, to add {\em burst} behavior that often occurs in practice, in simulating events at station $i$, with $5\%$ probability we pick a random integer $k$ between $1$ and $9$. Otherwise, for the other $95\%$, we set $k = 1$. We skip $k - 1$ segments of length $1/\lambda_i$ each and pack $k$ events in the next segment of length $1/\lambda_i$. Note that in terms of buses, this data generating process means that more than $20\%$ of buses come in short bursts. The computed average delay is given in Figure~\ref{figure:uniform-burst}, which yields a larger optimal $J_2$ but still smaller than that over Poisson processes. 

\subsection{Convergence of the Optimal Schedule}

We have shown that the variance of the fraction of observations at each station converges to zero at a particular rate as the number of cycles increases (Theorem~\ref{theorem:convergence}). As noted there, the optimal policy is such that the same convergence rate was observed at each station. In other words, the optimal policy not only balances the fraction of observations at each station, but also balances the convergence rates.

\begin{figure}[htp]
\begin{center}
    \includegraphics[width=2.8in]{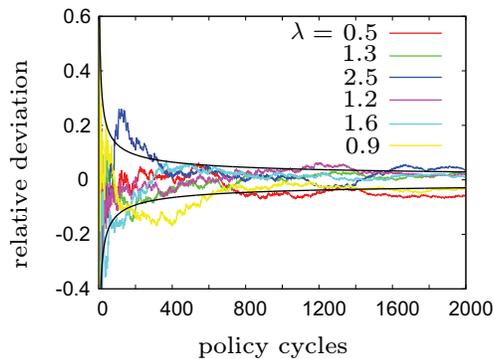} 
\end{center}
\caption{\label{figure_convergence_rate} The relative deviations of the data collecting process at the six stations with different $\lambda_i$'s over $2000$ policy cycles. The two black lines are computed with Theorem~\ref{theorem:convergence}.}
\end{figure}

Figure~\ref{figure_convergence_rate} depicts this phenomenon for a single execution of the optimal schedule for 2000 cycles. It is seen that the standard deviation converges to zero roughly with the rate computed in Theorem~\ref{theorem:convergence}. Moreover, the standard deviations are roughly the same across all stations.

\subsection{Robustness of Optimal Policies} 

\begin{figure}[htp]
\begin{center}
    \includegraphics[width=3in]{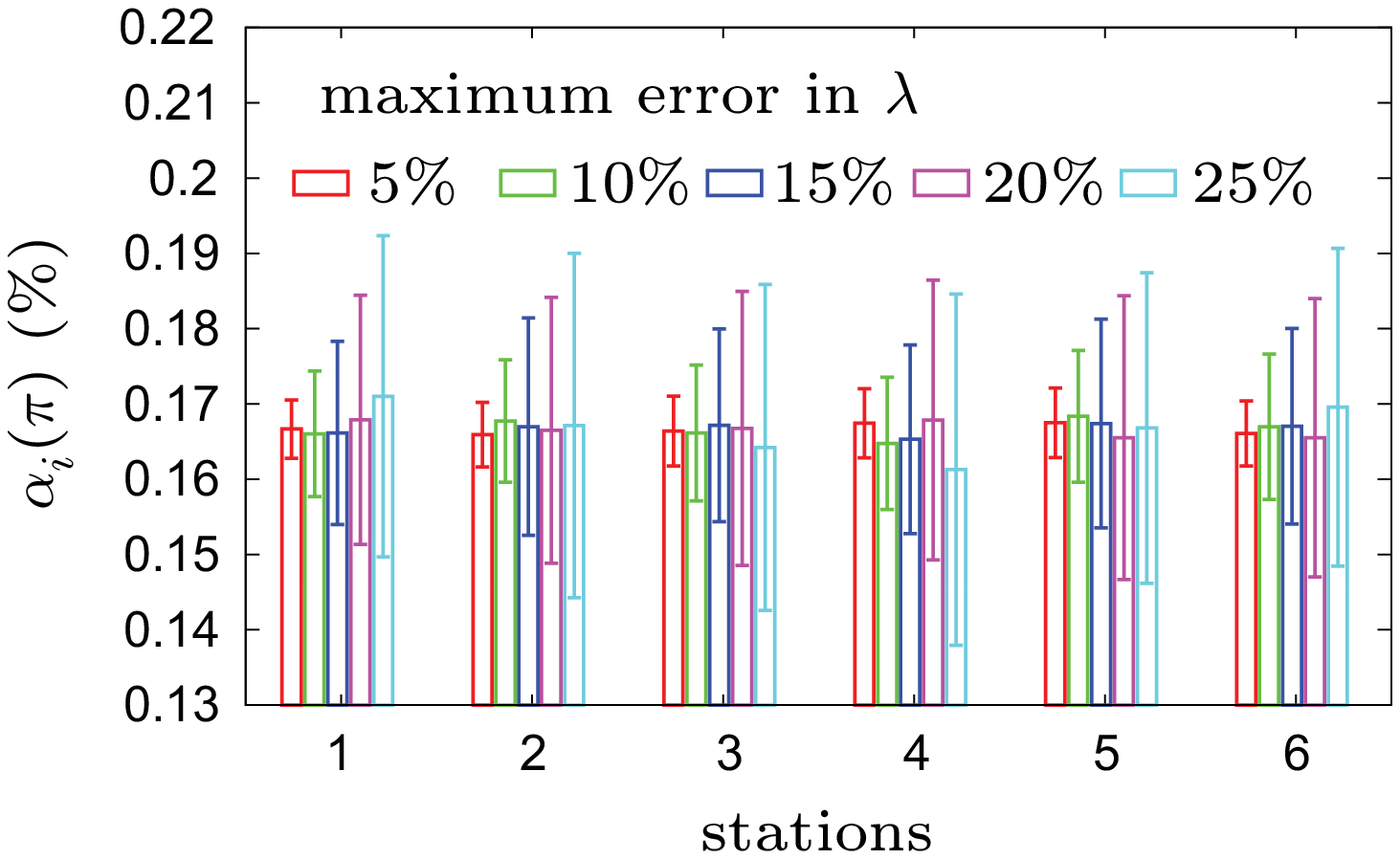} 
\end{center}
\caption{\label{figure_ratio_error} Simulated $\alpha_i(\pi)$ when running the optimal policy $\pi$ in environments with uncertainties in $\lambda_i$'s.}
\end{figure}

With Theorem~\ref{theorem:robustness}, one can expect the optimal policy to be robust in the sense that small estimation errors in the arrival rates should not greatly affect the performance of an optimal policy. 
\begin{figure}[htp]
\begin{center}
    \includegraphics[width=3in]{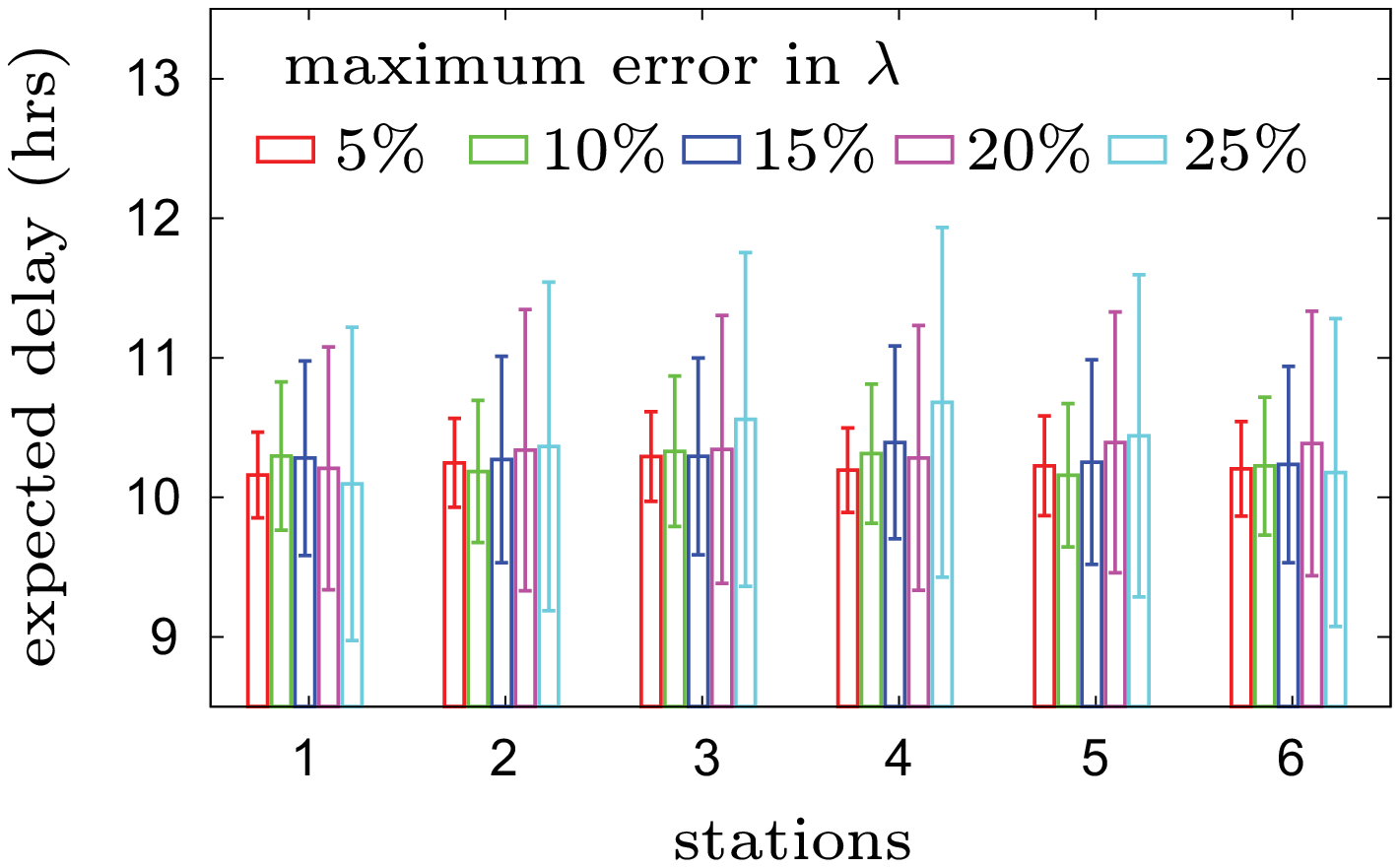} 
\end{center}
\caption{\label{figure_delay_error} Simulated delays ($\EEE[T_i(\pi)]$) when running the optimal policy $\pi$ in environments with uncertainties in $\lambda_i$'s.}
\end{figure}
We now use simulation to illustrate the robustness of an optimal policy. In our simulation based on the same $\lambda_i$'s (note that in this case, $\lambda_i/\sigma > 2$ holds for all $i$'s), we assume that the actual event arrival rate may vary up to $25\%$ (assuming randomly distributed errors in $\lambda_i$'s). For each error threshold from $5\%$ to $25\%$, 100 simulations were performed using environments based on these random (fixed) $\lambda_i$'s, over which the same optimal policy was ran for $10000$ policy cycles. The results were plotted in Figure~\ref{figure_delay_error} and Figure~\ref{figure_ratio_error}.

Figure~\ref{figure_delay_error} shows that using the same policy, one can expect relatively stable performance despite fairly large error in the estimated $\lambda_i$'s. For example, with up to $25\%$ maximum error, $\EEE[T_i(\pi)]$ only varies about $10\%$ across all stations at one standard deviation ({\em i.e.}, it is not very sensitive to the magnitude of $\lambda_i$'s). Similar behavior can be observed for $\alpha_i(\pi)$: up to $25\%$ error in $\lambda_i$'s yields a standard deviation of about $25\%$ in $\alpha_i(\pi)$ across all stations. 

\begin{figure}[htp]
\begin{center}
\begin{tabular}{cc}
    \includegraphics[width=1.4in]{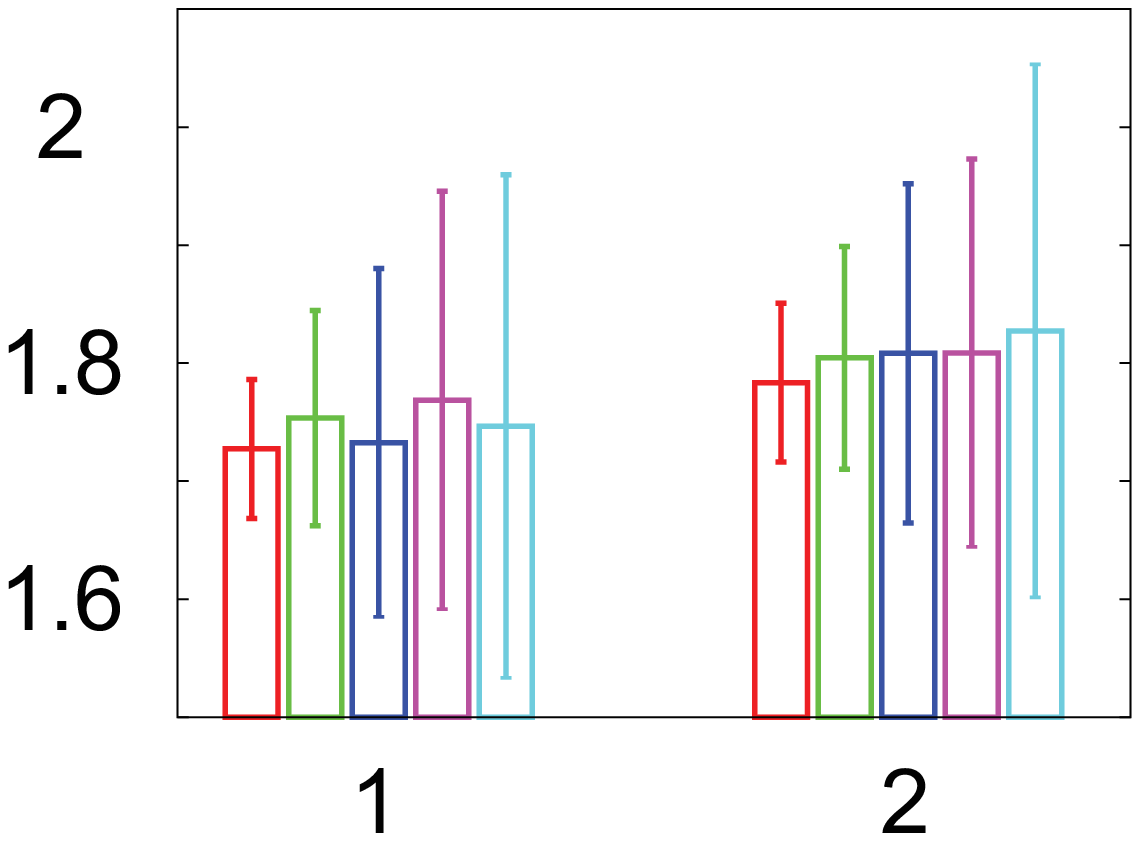} &
    \includegraphics[width=1.4in]{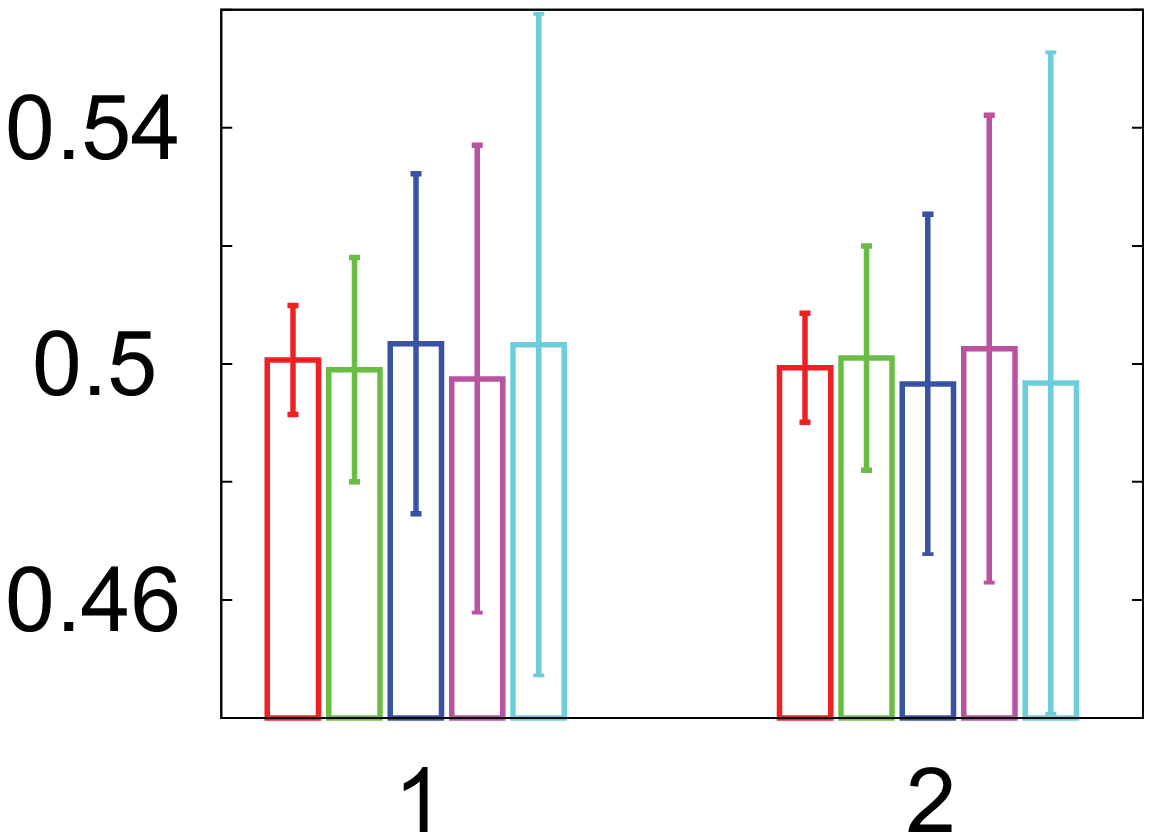} \\
    (a) & (b)
\end{tabular}
\end{center}
\caption{\label{figure_extreme_error} Simulated $\EEE[T_i(\pi)]$ and $\alpha_i(\pi)$ when running the optimal policy $\pi$ in environments with uncertainties in $\lambda_i$'s. Here The policy is generated based on $\lambda_1 = 1$ and $\lambda_2 = 100$. The two graphs correspond to Figure~\ref{figure_delay_error} and Figure~\ref{figure_ratio_error}, respectively. We omitted axes labels and legends that have identical meanings with those in Figure~\ref{figure_delay_error} and Figure~\ref{figure_ratio_error}.}
\end{figure}

Though not directly implied by Theorem~\ref{theorem:robustness}, an optimal policy also appears to be stable with respect to widely varying stochastic arrival rates. Taking an extreme example having two stations with $\lambda_1 = 1$, $\lambda_2 = 100$ (here $\lambda_2/\sigma = 1.01 < 2$), and $\tau_{12} = \tau_{21} = 0.1$, we performed the same experiments on stability, the results of which are captured in Figure~\ref{figure_extreme_error}. The deviations are similar to what we observed in Figure~\ref{figure_delay_error} and Figure~\ref{figure_ratio_error}. The optimal policy here is $\pi = (0.5702, 0.0057)$. 

\section{Conclusion}\label{section:conclusion}
We introduced a novel persistent monitoring and data collection problem in which transient events at multiple stations arrive following stochastic processes. We studied the performance of cyclic policies on two objectives: {\em (i)} maximizing the minimum fraction of expected events to be collected at each station so that no station receives insufficient or excessive monitoring effort, and {\em (ii)} minimizing the maximum delay in observing two consecutive events generated by the same process between policy cycles. We focused on an important case in which the locations to be visited form a closed chain. We showed that such a problem admits a (often unique) cyclic policy that optimizes both objectives. We also showed that the second, more complex objective function is quasi-convex, allowing efficient computation of the optimal policy with standard gradient descent methods when the cyclic ordering of the stations is fixed. Moreover, our study important properties of the optimal solution, including convergence rate and robustness result, further offered us insights that lead to an polynomial-time approximation algorithm for the more general problem in which the cyclic order is unknown {\em a priori}. 

\begin{figure}[htp]
\begin{center}
    \includegraphics[width=2.8in]{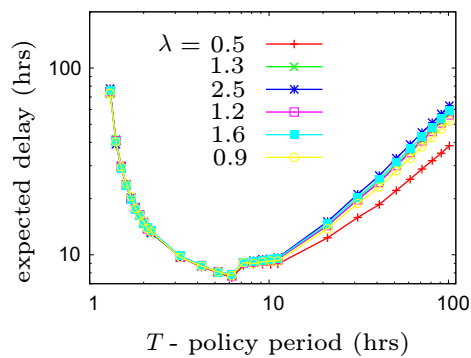} 
\end{center}
\caption{\label{figure:delay-feedback} Simulated average delay with feedback such that the robot will leave a station once the expected number of events per cycle is collected. The corresponding expected delays without feedback are plotted in Figure~\ref{fig:exp_delay_more_zoom}.}
\end{figure}

Our study also raises many interesting and well formulated open problems; we discuss two here. First, in our formulation, the robot is not required to process the data it collects while waiting at the stations. Whereas this assumption applies to many scenarios, it is perhaps equally natural to assume the opposite and let the robot know when it observes an event. This then gives rise to feedback or adaptive policies. For example, one way to design such a policy is to let the robot move away from a station once it knows enough number of events have been collected at the station. Intuitively, such feedback policies should do better due to the memoryless property of Poisson process. Preliminary simulation result confirms our hypothesis (see Figure~\ref{figure:delay-feedback}). Interestingly, such feedback policies seem to induce discrete jumps in the average delay, which we look forward to understanding in future research. Another interesting related angle is to allow sensors to have non-trivial footprint. That is, the mobile sensor is able to cover multiple stations simultaneously. 

Second, we had initially conjectured that TSP-based cyclic policies might be the best policies for the proposed multi-objective optimization problem without requiring the cyclic policy assumption. This turns out not to be the case; Appendix~\ref{appendix:example} provides a counterexample. In the counterexample, a {\em periodic policy}\footnote{In contrast to a cyclic policy, which allows a single visit to each station during a policy period, a periodic policy allows multiple visits to the same station during a single policy period.} is proven to be better than the TSP-based cyclic policy. This observation prompts at least two open questions: {\em (i)} how we may find the optimal periodic policy for the proposed multi-objective optimization problem? {\em (ii)} are periodic policies the best policies without feedback? 
\bibliographystyle{IEEEtranN}
\bibliography{references}

\appendix
\subsection{Technical Proofs}\label{appendix:proofs}
\noindent{\sc Proof of Lemma \ref{lemma:quasi-convexity}.} For notational convenience, define $\gamma_i :=  \sigma / \lambda_i$. Note that we implicitly use the fact that all functions used in the proof are continuous. Substituting $\tobs = T - \ttr$ and $t_i = \gamma_i \tobs$ into the RHS of~\eqref{eq:exp_break_cyclic} yields
$$
\begin{array}{l}
\EEE[T_i(\pi)] = \displaystyle\frac{2}{\lambda_i} + \frac{T - t_i - (T-t_i)e^{-\lambda_it_i} + (T - 2t_i)e^{-\lambda_it_i}}{1-e^{-\lambda_it_i}}\\
\quad =\displaystyle\frac{2}{\lambda_i} + \frac{T - t_i - t_ie^{-\lambda_it_i}}{1-e^{-\lambda_it_i}}\\
\quad =\displaystyle\frac{2}{\lambda_i} + \frac{\tobs + \ttr   - \gamma_i\tobs - \gamma_i \tobs e^{-\lambda_i\gamma_i\tobs}}{1-e^{-\lambda_i\gamma_i\tobs}}.
\end{array}
$$

Noting that by scaling the unit of time, we may assume that $\lambda_i = 1$. Using this and letting $x := \gamma_i \tobs$ gives us
$$
\begin{array}{l}
\EEE[T_i(\pi)] = \displaystyle 2 + \frac{\ttr  + (\frac{1}{\gamma_i}  - 1)x - xe^{-x}}{1-e^{-x}} \\
\quad = \displaystyle 2 + \frac{\ttr  + (\frac{1}{\gamma_i}  - 2)x}{1-e^{-x}} + x,
\end{array}
$$
in which $\ttr  > 0$ and $\gamma_i \in (0, 1)$. For convenience, we let $\alpha := \ttr$ and $\beta = 1/\gamma_i - 2$.  Showing that $\EEE[T_i(\pi)]$ is quasi-convex is equivalent to showing that 

$$
\begin{array}{l}
f(x) := \displaystyle \frac{\alpha + \beta x}{1-e^{-x}} + x
\end{array}
$$
is quasi-convex for $x > 0$,\footnote{In the rest of the proof, unless explicitly stated otherwise, the domain of $x$ is assumed to be $(0, \infty)$.} $\alpha > 0$, and $\beta > -1$, the second derivative of which is
$$
\begin{array}{l}
f''(x) = \displaystyle \frac{e^x(\alpha(e^x + 1) + \beta (e^x(x -2) + x + 2))}{(-1+e^{x})^3}.
\end{array}
$$

Since $e^x(x -2) + x + 2$ is strictly positive,\footnote{To see this, let $h(x) = e^x(x -2) + x + 2$; then $h(0) = 0$, $h'(0) = 0$, and $h''(x) = xe^x > 0$ for all $x > 0$. Therefore, $h'(x) > 0$ and $h(x) > 0$ for all $x > 0$.} $f''(x) > 0$ for $\beta \ge 0$. Therefore, $f(x)$ is convex for $\beta \ge 0$. We are left to show that $f(x)$ is quasi-convex for $\beta \in (-1, 0)$. We proceed by first establishing some properties of the function
$$
\begin{array}{l}
g(x) = \displaystyle \alpha(e^x + 1) + \beta (e^x(x -2) + x + 2)
\end{array}
$$
for $\alpha > 0$, and $\beta \in (-1, 0)$. We have $g(x) \in C^{\infty}$ for $x \ge 0$, $g(0) = 2\alpha > 0$, $\lim_{x\to \infty}g(x) = - \infty$, 
$$
\begin{array}{l}
g'(x) = \displaystyle (\alpha + \beta x - \beta)e^x + \beta,
\end{array}
$$
and
$$
\begin{array}{l}
g''(x) = \displaystyle (\alpha + \beta x)e^x.
\end{array}
$$

Because $(\alpha + \beta x)$ is linear, monotonically decreasing and crosses zero at most once, and $e^x$ is positive and strictly increasing, $g''(x)$ has at most a single local extrema (a maxima) before it crosses zero. Therefore, $g'(x)$ has at most two zeros and must first increase monotonically and then decrease monotonically, implying that $g(x)$ has at most three zeros. Since $g(0) > 0$ and $\lim_{x\to \infty}g(x) = - \infty < 0$, $g(x)$ has either one or three (but not two) zeros. For $g(x)$ to have three zeros, $g'(x)$ must have two zeros. Since $\lim_{x\to \infty}g'(x) = -\infty$ (because $\beta x e^x$ eventually dominates and $\beta < 0$), we must have $g'(0) < 0$. This is not possible because $g'(0) = \alpha > 0$. Therefore, $g'(x)$ can cross zero and change sign at most once,\footnote{Alternatively, solving $g'(x) = 0$ in Mathematica yields at most a single zero in $(0, \infty)$ at 
$x = \displaystyle \frac{\beta W(-e^{\frac{\alpha}{\beta}-1}) -\alpha + \beta}{\beta}
$, in which $W(\cdot)$ is the (principal) {\em Lambert W-function}.} implying that $g(x)$ has a single zero. That is, $g(x)$ is positive for small $x$ and then remains negative after crossing zero. Because
$$
f''(x) = \displaystyle \frac{e^x g(x)}{(-1+e^{x})^3}
$$
and $e^x/(-1+e^{x})^3$ is strictly positive, $f''(x)$ behaves similarly as $g(x)$ ({\em i.e.}, $f''(0) > 0$, crosses zero only once as $x$ increases, and stays negative after that). This implies that for every fixed $\alpha > 0$ and $\beta \in (-1, 0)$, there exists $x_0 > 0$ such that $f(x)$ is convex on $x \in (0, x_0)$ and concave on $x \in (x_0, \infty)$. Now because $f(x) \to \infty$ for both $x \to 0^+$ and $x \to \infty$, and $f(1) < \infty$, $f(x)$ must have a single local minima (and therefore, a single global minima on $\mathbb R^+$). To see that this is the case, as $f(x)$ turns from convex to concave at $x = x_0$, we must have $f'(x_0) \ge 0$ because otherwise $f'(x) < 0$ for $x > x_0$ due to $f(x)$'s concavity. We then have $\lim_{x\to \infty}f(x) < \infty$, a contradiction. Thus, $f(x)$ has a single minimum on $x \in (0, x_0)$. Finally, to see that $f(x)$ is quasi-convex, we note that $\lim_{x\to \infty}f'(x) = 1 + \beta > 0$, implying that $f'(x) > 0$ on all $x \in (x_0, \infty)$. We then have that $f(x)$ is monotonically increasing on $x \in (x_0, \infty)$. From here, the quasi-convexity of $f(x)$ can be easily shown following definitions. ~\qed 

\subsection{Non-Optimal TSP Cyclic Policies}\label{appendix:example}
In this part of the appendix, we provide an example problem for which the optimal TSP cyclic policy is not the optimal policy for maximizing $J_1$ and minimizing $J_2$. We build the problem in two steps. Our initial problem, which is to be updated in a little while, has three stations with input parameters
$$
\lambda_1 = \lambda_3 = 1, \lambda_2 = 2, \tau_{1,2} = \tau_{2,3} = 0.1, \tau_{3,1} = 0.2.
$$ 


Using Algorithm 1, we compute the optimal cyclic policy as $\pi_1 = (t_1 = 0.53, t_2 = 0.27, t_3 = 0.53)$, which contains a TSP tour of the stations. We may further compute $\EEE[T_1(\pi_1)] = \EEE[T_3(\pi_1)] = 4.15$ and $\EEE[T_2(\pi_1)] = 4.17$, which implies that $J_2(\pi_1) = 4.17$. Then, we modify $\pi_1$ to get another policy $\pi_2$, which is a periodic policy, by changing the visiting order of the stations to $1, 2, 3, 2, 1, \ldots$, {\em i.e.}, station 2 is visited twice as frequently, and letting the robot stay at station 2 for $t_2/2$ time for each visit. Employing the proof technique of Lemma~\ref{l:exp_break_cyclic}, we compute that $\EEE[T_2(\pi_2)] = 3.26$ whereas $\EEE[T_i(\pi_2)] = \EEE[T_i(\pi_1)] = 4.15$ for $i \in \{1, 3\}$. Thus $J_2(\pi_2) = 4.15$. Because $J_1(\pi_1) = J_1(\pi_2) = 1/3$, $\pi_2$ is a better periodic policy than $\pi_1$. 

We now construct the final example problem with again three stations and update the parameters to
$$
\lambda_1 = \lambda_3 = 1, \lambda_2 = 2, \tau_{1,2} = \tau_{2,3} = 0.1 + \epsilon, \tau_{3,1} = 0.2,
$$ 
in which $\epsilon > 0$ is a small perturbation. That is, we make the two paths between stations $i$ and $i+1$, $i \in \{1, 2\}$, a little longer. As long as $\epsilon > 0$, we have that a robot executing $\pi_2$ will travel a strictly longer distance during each policy period than a robot executing $\pi_1$, the TSP-based cyclic policy. However, by continuity, for a small enough $\epsilon$, $\pi_2$ will remain a better policy than $\pi_1$ for optimizing $J_1$ and $J_2$.

\end{document}